\RequirePackage{amsthm}
\documentclass[sn-aps,Numbered,iicol]{sn-jnl}

\usepackage{graphicx}%
\usepackage{multirow}%
\usepackage{amsmath,amssymb,amsfonts}%
\usepackage{mathrsfs}%
\usepackage[title]{appendix}%
\usepackage{xcolor}%
\usepackage{textcomp}%
\usepackage{manyfoot}%
\usepackage{booktabs}%
\usepackage{algorithm}%
\usepackage{algorithmicx}%
\usepackage{algpseudocode}%
\usepackage{listings}%
\usepackage{subfigure}
\newcommand{\ts}{\textsuperscript}

\theoremstyle{thmstyleone}%
\theoremstyle{thmstyletwo}%

\theoremstyle{thmstylethree}%

\begin{document}

\title[Machine Learning Based Complex Information Extraction from Historical Documents]{Improving OCR Quality in 19\ts{th} Century Historical Documents Using a Combined Machine Learning Based Approach}


\author*[1]{\fnm{David} \sur{Fleischhacker}}\email{david.fleischhacker@student.tugraz.at}

\author[1,2]{\fnm{Roman} \sur{Kern}}\email{rkern@tugraz.at}
\equalcont{These authors contributed equally to this work.}

\author[3]{\fnm{Wolfgang} \sur{G\"oderle}}\email{wolfgang.goederle@uni-graz.at}
\equalcont{These authors contributed equally to this work.}

\affil*[1]{\orgdiv{Institute of Interactive Systems and Data Science}, \orgname{Graz University of Technology}, \orgaddress{\street{Sandgasse 36}, \city{Graz}, \postcode{8010}, \country{Austria}}}

\affil[2]{\orgname{Know Center}, \orgaddress{\street{Sandgasse 36}, \city{Graz}, \postcode{8010}, \country{Austria}}}

\affil[3]{\orgdiv{History Department}, \orgname{University of Graz}, \orgaddress{\street{Attemsgasse 8}, \city{Graz}, \postcode{8010}, 
\country{Austria}}}


\abstract{This paper addresses a major challenge to historical and fellow humanist research with regard to the long 19\ts{th} century, as large quantities of sources, frequently printed, have become available for the first time, while extraction techniques are lagging behind.
Research has so far neglected the vast majority of these printed serial sources, with the exception of a few high-profile data series, such as census data, which have mainly been laboriously extracted manually. 
Therefore, we researched machine learning (ML) models to recognise and extract complex data structures in a high-value historical primary source, Habsburg Central Europe's \textit{Hof- und Staatsschematismus}, which was published between 1702 and 1918. 
It records every single person in the Habsburg civil service above a certain hierarchical level and documents the genesis of the central administration over two centuries. Its complex and intricate structure as well as its enormous size have so far made any more comprehensive analysis of the administrative and social structure of the later Habsburg Empire on the basis of this source impossible. 
\\
We pursued two central objectives: Primarily, the improvement of the OCR quality, for which we considered an improved structure recognition to be essential; in the further course, it turned out that this also made the extraction of the data structure possible. 
We chose Faster R-CNN as base for the ML architecture for structure recognition. 
In order to obtain the required amount of training data quickly and economically, we synthesised \textit{Hof- und Staatsschematismus}-style data, which we used to train our model. The model was then fine-tuned with a smaller set of manually annotated historical source data. We then used Tesseract-OCR, which was further optimised for the style of our documents, to complete the combined structure extraction and OCR process. 
\\
We then evaluated our model's performance in two steps, first we compared our fine-tuned Tesseract's performance with that of an off-the-shelf version of Tesseract, then we combined the structure recognition and the fine-tuned Tesseract and evaluated the results against that of a standard Tesseract distribution. Results show a significant decrease in the two standard parameters of OCR-performance, WER and CER (where lower values are better), already in the first test, when the refined model fared by 64.24\% better in terms of the CER and by 40.91\% when it comes to the WER. Combined structure detection and fine-tuned OCR improved CER and WER values by remarkable 71.98\% (CER) respectively 52.49\% (WER). 
These results highlight the usefulness of ML techniques for extraction of historical documents and to aid research to ease the extraction of information from historical documents for further analysis.}




\maketitle

\section{Introduction}\label{sec1}

The long 19\ts{th} century provides historians and fellow humanists with a wealth of sources, handwritten and printed alike. So far only a small fraction of these sources have been made accessible and considered for research. Particularly the rich collection and publication of relevant information in State Manuals, which can be observed from the early 18\ts{th} century onward, represents a bulk of primary sources of immense value. In the case of Habsburg Central Europe, the so-called \textit{Hof- und Staatsschematismus}, which was published from 1702 to 1918, provides us with a set of serial data on the administrative and representative elites of Habsburg Central Europe in very high quality. Although this information is not published continuously – there are several large gaps in between – a thorough analysis of this data set would contribute decisively to a significantly better understanding of relevant social processes, power dynamics, social networks and careers in modern Central Europe. The Schematismus could be used to trace the genesis of state and administrative institutions, their functioning and development, and the professional biographies of tens of thousands of officials and decision-makers over more than two centuries, across political ruptures and social transformations.

However, the complex structure of such publications has made a more comprehensive and quantitative evaluation of these sources impossible. These manuals are usually available in printed form. Even though OCR-quality has improved dramatically since the early days of the retro-digitisation of historical publications, structure analysis and layout detection remain a problem. Particularly the multi-column layouts, deeply branched hierarchies of several levels, and multimodal page designs that include text next to tables and next to complex lists that can be encountered in the \textit{Schematismus} are representative of the challenges that digital historians and humanists are confronted with when they have to process large quantities of such documents. Thus, there has not yet been a comprehensive extraction of information, entire data sets or structures, such as information on hierarchies, on the detailed composition of administrative authorities, or simply careers or biographies with regard to the \textit{Schematismus} yet. Some studies manually extracted relevant information, which proved tedious, time-consuming and prone to errors. 

The Habsburg Court and State Manuals are considered a highly valuable source. For years consideration has been given to publishing parts of them as digital editions. Such undertakings have so far reliably failed because of the immense size of the task; one of the more popular series of such handbooks, published by the Austrian National Library\footnote{\url{alex.onb.ac.at}}, comprises 153 volumes compiled between 1702 and 1918, and is estimated to contain between 110,000 and 150,000 printed pages. While in theory indexing via OCR and enabling a keyword search seems at least conceivable, in practice it has also been shown that OCR recognition primarily of proper names (which are likely to account for a mass share of the total text of well over 60 percent; if offices and institutions, whose names are usually not listed in the relevant thesauri either, are also included, the share is again significantly higher) produces a significantly higher error rate than for OCR in the standard text. 

In a first step we aim to address and improve the OCR quality.
In a second step we reached out to address the issue of machine learning driven information extraction from a deeply locked-in textual structure. The solution we present here will serve as a first step toward enabling a large scale extraction of information from State Manuals to feed the data into a research database that will allow historians and fellow humanists to execute complex queries and to address an array of new research questions in order to produce a more comprehensive and detailed understanding of Central European social elites in the long 19\ts{th} century.




\section{Background \& Related Work}\label{sec3}

Wide variation in document formats and degradation over time make automated layout detection quite challenging, especially for historical documents. For these reasons, a broad range of techniques and methods has been developed and applied over time, in order to be able to provide for a better automated information extraction from historical documents.\\ 
In this section, we will first present the state of research in history and more broadly the humanities, then the key layout detection methods that have been developed and used in the field of historical document analysis are discussed. A greater emphasis will be placed on Convolutional Neural Networks, especially faster R-CNNs and mask R-CNNs, as these deep learning models later turned out to be particularly well-fitted to tackle the task of layout detection with regard to the \textit{Hof- und Staatshandbücher}. 

\subsection{Layout detection and OCR of historical documents}

The retro-digitisation of either parts of or entire historical sources has been an issue among historians and fellow humanists since at least the 1950s, particularly with regard to serial sources \citep{Raphael1994}. However, for the larger part of the past seven decades, information extraction and the production of digital data that could be processed by computing machines, has been executed manually. Even though OCR software has been widely available by the 1990s at latest, particularly processing historical data has remained a complex issue\citep{Jannidis2017}, as the quality of results has been varying strongly \citep{Boros2022}. We therefore observe a bifurcation in the field of digital historical and humanist research: Whereas large amounts of retro-digitised historical data are processed automatically by many important providers of research data, such as www.archive.org \citep{Wajer2021}, many national libraries, as for instance the Austrian National Library \citep{ABO2023}, the National Library of Finland \citep{Kettunen2020} or the Munich Digitization Center of the Bavarian State Library \citep{MDZ2023}, and large transnational initiatives, as for example Europeana \citep{Markus2019}, users of these data, such as historical research projects, are still relying on the manual transcription or annotation of digitised primary sources, even at scale.\footnote{There is very little documentation on how data in historical research projects is OCRed, so we have to refer to information that we obtained in personal conversation with several dozen colleagues over the past one-and-a-half decade. Based on the information available to us, we would assume that manual transcription of historical text, but especially tabular data, is still the standard procedure, even though this usually starts with a raw document created with standard OCR (Tesseract or Transkribus), which is subsequently improved. The digitisation of tabular data is particularly challenging because the target structure usually has to be created manually beforehand. Cf. \citep{Jannidis2017}} Relevant data bases that were build predominantly on data extracted by manual labour include the \textit{Wiener Datenbank zur Europäischen Familiengeschichte} \citep{Ehmer}, \textit{The Emperor's Desk} \citep{Becker}, the prosopographical data processed by \textit{The Viennese Court} \citep{Romberg}, and further the projects run in \textit{Social Mobility of Elites} \citep{Popovici}.
\\ 
Historians and digital humanists working with retro-digitised text data are familiar with the phenomenon that OCR-ed texts on a platform may vary in quality, mostly due to the fact that texts underwent OCR procedure at different times and different technologies were used. As a result, full-text searches sometimes produce inhomogeneous results. Also, a "systematic" OCR error, i.e. distortions that are typical for certain OCR software, can no longer be processed in such a targeted manner if large text corpora consist of parts that were OCRed at different times with different software. In many projects, therefore, raw digitised documents are now re-OCRed, although many historical layouts still pose a challenge for standard OCR but also specialised software \citep{Kettunen2020}.\footnote{Details about the OCR process at large providers of cultural heritage data are only available to a very limited extent, and the publicly accessible documentation leaves much to be desired, even at public institutions.} Apart from specialized solutions \citep{Engl2020}, Tesseract OCR and Transkribus are currently regarded as reference standards in the field of historical OCR, but still encounter limitations that require a high level of manual effort, given the special layouts and structures that will be discussed here \citep{Martinek2020}. Libraries have begun to use the potential of ML to the classification of large quantities of texts \citep{Cordell2020}, especially in the context of research libraries \citep{Gasparini_Kautonen_2022}.
\\
For historical research, primary sources such as censuses of the Habsburg Empire have been mostly digitized manually \citep{Teibenbacher2012} \citep{Zechner2021}. However, because of the enormous volume, this method is not feasible for other complex source works, such as the \textit{Schematismus}. Even in the last initiative we know of, the size and complexity of the \textit{Schematismus} was ultimately considered insurmountable for only partially automated data extraction, and manual edition of a small part was envisaged as an alternative. Due to these obstacles, very little research is engaging with a deeper exploration of the information stored in the \textit{Schematismus}, the work of Bavouzet \citep{Bavouzet2019} (building entirely on manually extracted data) is clearly standing out as a beacon here. 

\subsection{Rule-based methods, SVMs and Clustering}\label{subsec1}

Rule-based approaches to detecting the layout of documents have been around since the early 90s \citep{118166}. The basic procedure here is to split layout detection into two parts. In the first part, the documents are segmented into text and non-text areas. This is accomplished by performing a four-step algorithm called the Run Length Smoothing Algorithm (RLSA). A RLSA image is produced by first smoothing document images horizontally and vertically, then applying an AND-operation on both results, and then smoothing them again horizontally. Smoothing in both horizontal and vertical directions is performed by calculating the distance between two adjacent black pixels. When this distance is less than a predefined threshold, both pixels are considered joined, and all white pixels in between will be turned black. Once such a RLSA image has been produced, boundaries can be established around the smoothed areas by applying connected component analysis \citep{9112}. After calculating various statistics on these individual areas, such as black pixel density, aspect ratio or height, the first step in detecting the layout has been completed. The second part of this rule-based approach is to actually use rules to classify each boundary box into three categories: text, non-text and unknown. The final classification is made by comparing the calculated statistics of each box with predefined parameters that were calculated based on data that was known to belong to one of the categories.

There are, however, problems associated with classifying blocks of interest according to fixed layout rules. As the general structure of a document can differ drastically between different types of documents, it is possible that they can differ even within the same type of document. For instance, the title page of a newspaper may have a completely different structure than any other page. As a result, the rule-based layout detection approach has been further fine-tuned by using knowledge of the type of document in question \citep{599038}. Some global rules will apply to the majority of documents, while others will be specific to the type of document being analysed. In particular, this knowledge-based approach is made up of three levels of rules: knowledge rules, control rules, and strategy rules. A knowledge rule defines some common characteristics and spatial constraints for different document blocks. Control rules, on the other hand, regulate the analysis of the image, determine when consistency has been achieved, and then invoke appropriate knowledge rules. Lastly, strategy rules determine when control rules are invoked. Using this knowledge-based layout detection approach, a tree of classified blocks in the document image is produced along with all the relevant extracted feature details.

However, there are unresolved challenges with regards to rule-based layout detection of historical documents. 
Every document type requires predefined rules in order to accurately detect layout. This is especially challenging with historical documents, since the structure of documents varies greatly from one time period and location to another. Moreover, in most cases these predefined rules are derived from template documents, which can be very time- and labour-intensive. Consequently, rule-based layout detection algorithms are limited in their ability to detect unexplored layouts, making them less suitable as a layout detection method for a wide variety of historical documents.
Thus, traditional machine learning approaches have gained increasing popularity as a method of detecting layouts in historical documents. For instance, Support Vector Machines (SVM) were used to first classify the text in scanned documents. Then these blobs of text were clustered into paragraphs \citep{diem2011text}. This method was successfully used to analyse and reconstruct documents from over 600 million snippets of Stasi files that were discovered after the fall of the Berlin Wall.

Following the classification of text blobs, text clustering is achieved by taking the minimum area rectangle for each word blob and extending its major axis by a so- called fuse factor. Using the remaining minimum area rectangles, a fusion test is conducted to group words within text lines and paragraphs. By using the weight histograms of the grouped children, the class labels are re-computed once the words are grouped. Voting against the weights of the children is employed to correct falsely classified words in text classification by back propagation. Rather than relying on strict class decisions, this method incorporates a global class decision and propagates weights to improve classification performance.
Based on the data set used in the ICDAR2009 Page Segmentation Competition \citep{antonacopoulos2009icdar}, the proposed method beat the then state-of-the-art methods in the overall page segmentation task with an F-score of 0.945. 

\subsection{CNNs}\label{subsec2}
\begin{figure}[ht]%
\centering
\includegraphics[width=\columnwidth]{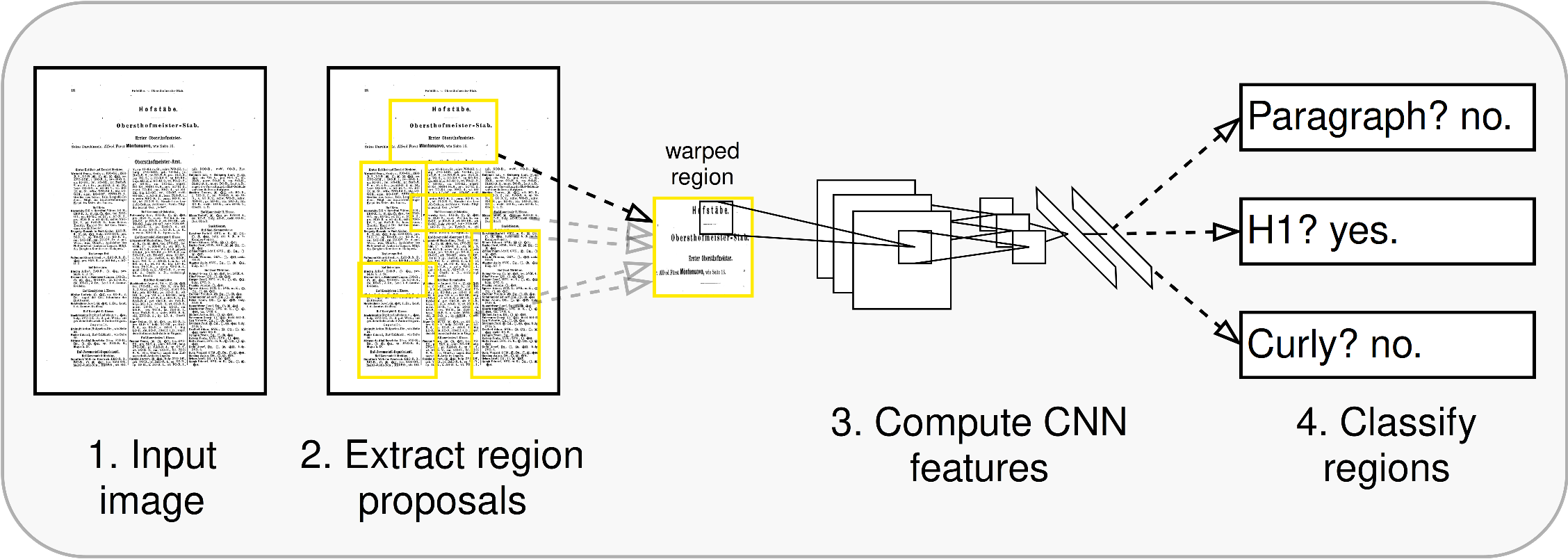}
\caption[R-CNN pipeline.]{An example of the R-CNN pipeline: starting with the input image, region proposals are generated and passed through a CNN to extract features, which are then used to classify and localize objects \citep{6909475}.}
\label{fig:r-cnn}
\end{figure}

Since the rise in popularity of Artificial Neural Networks (ANN) over the last few years, convolutional neural networks (CNN) have gained massive attention in image driven pattern recognition \citep{https://doi.org/10.48550/arxiv.1511.08458}. This type of neural network is specifically designed to work with image data. Therefore, convolutional neural networks have also sparked interest in the field of historic document analysis. During the 22\ts{nd} International Conference on Pattern Recognition 2014, convolutional neural networks were found to be as effective as state-of-the-art methods at the time in classifying document images \citep{kang2014convolutional}. Furthermore, the baseline detection competition at the International Conference on Document Analysis and Recognition (ICDAR2017) was dominated by convolutional neural networks and U-nets, which are similar to CNNs \citep{diem2017cbad}.\\
In general, a CNN consists of several layers, each composed of neurons \citep{8308186}. There are four main types of layers: initially, there is one called input, which represents the image input. Following are the convolutional- and pooling-layers. As they are invisible from the outside, they are often referred to as hidden layers. Convolutional layers produce activation maps by shifting filters (or kernels) over the input image, multiplying the kernel elements by the values of the input image, and finally summing up these multiplications. By contrast, pooling layers reduce the spatial dimensions by downsampling the output of the convolutional layers. Reducing the number of parameters in the network will assist in preventing overfitting and will allow the model to be more resilient to small changes in input, while also making training less resource-intensive. Fully connected layers, or dense layers, are the final layers. These layers are used to classify the features extracted from the convolutional layers. By interconnecting neurons in each layer with neurons in subsequent layers, the network can learn complex relationships between features. Finally, the fully connected layers produce a prediction based on the input data.\\
However, standard convolutional neural networks have difficulty detecting the layout of historical documents. Contrary to image classification, which assigns a single label based on the content of an image, object detection identifies various objects within an image. Elements such as headlines and paragraphs are categorised into predetermined groups and bounding boxes are drawn around them. 
Particularly in the case of document layout detection, these objects can appear multiple times on a single document page. Due to the fact that the output layer of a CNN has a fixed length, the amount of elements that can be detected at a time is limited by this size. To resolve this issue, the original input image needs to be divided into individual regions of interest. The convolutional neural network is then applied to each region separately. In the past, sliding windows of fixed sizes have been used to traverse the whole input image to get such regions \citep{655647}. This sliding-window approach, however, is very computationally expensive for high-resolution input images, like those found in digital documents. To that end, region proposals and convolutional neural networks have been combined to create an improved method known as R-CNN \citep{6909475}. Candidate regions are found within the input image using a selective search approach. An overview of the R-CNN method is illustrated in Figure \ref{fig:r-cnn}.

\begin{figure}[ht]%
\centering
\includegraphics[width=\columnwidth]{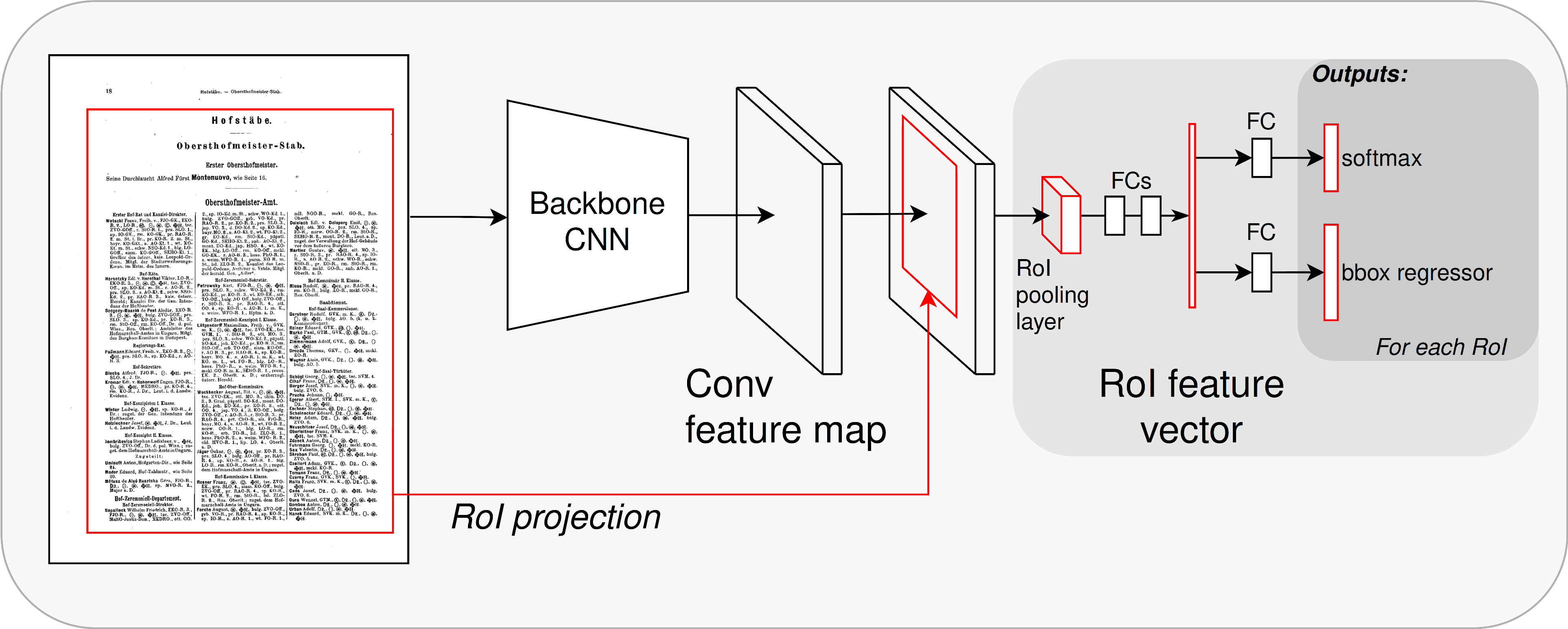}
    \caption[Fast R-CNN architecture.]{A visual representation of the Fast R-CNN architecture: starting from the input image, region proposals (RoI projection) are generated using the selective search algorithm and passed through a CNN to extract features, which are then used by a region of interest (RoI) pooling layer to extract a feature vector for each proposal. These feature vectors are then passed through twin layers of a softmax classifier and bounding box regression for the classification and localization of objects in the image \citep{fast-r-cnn}.}
    \label{fig:fast-r-cnn}
\end{figure}

Since R-CNN uses CNN as a feature extractor for each region proposal individually, evaluating a large number of candidate regions can be computationally intensive. As a result, a further improved method has been developed to overcome this drawback. This method is called fast R-CNN \citep{fast-r-cnn}. In principle, the fast R-CNN algorithm follows the same approach as the previously mentioned R-CNN algorithm. As opposed to applying the convolutional neural network to each region proposal individually, the entire input image is fed to a CNN to generate a convolutional feature map. Using this feature map, a fixed length feature vector is extracted for each object proposal. The feature vectors from each region of interest are fed into a sequence of fully connected layers to produce two outputs: a class prediction and a bounding box. Figure \ref{fig:fast-r-cnn} gives an overview of such a fast R-CNN pipeline.

\begin{figure}[ht]%
\centering
\includegraphics[width=\columnwidth]{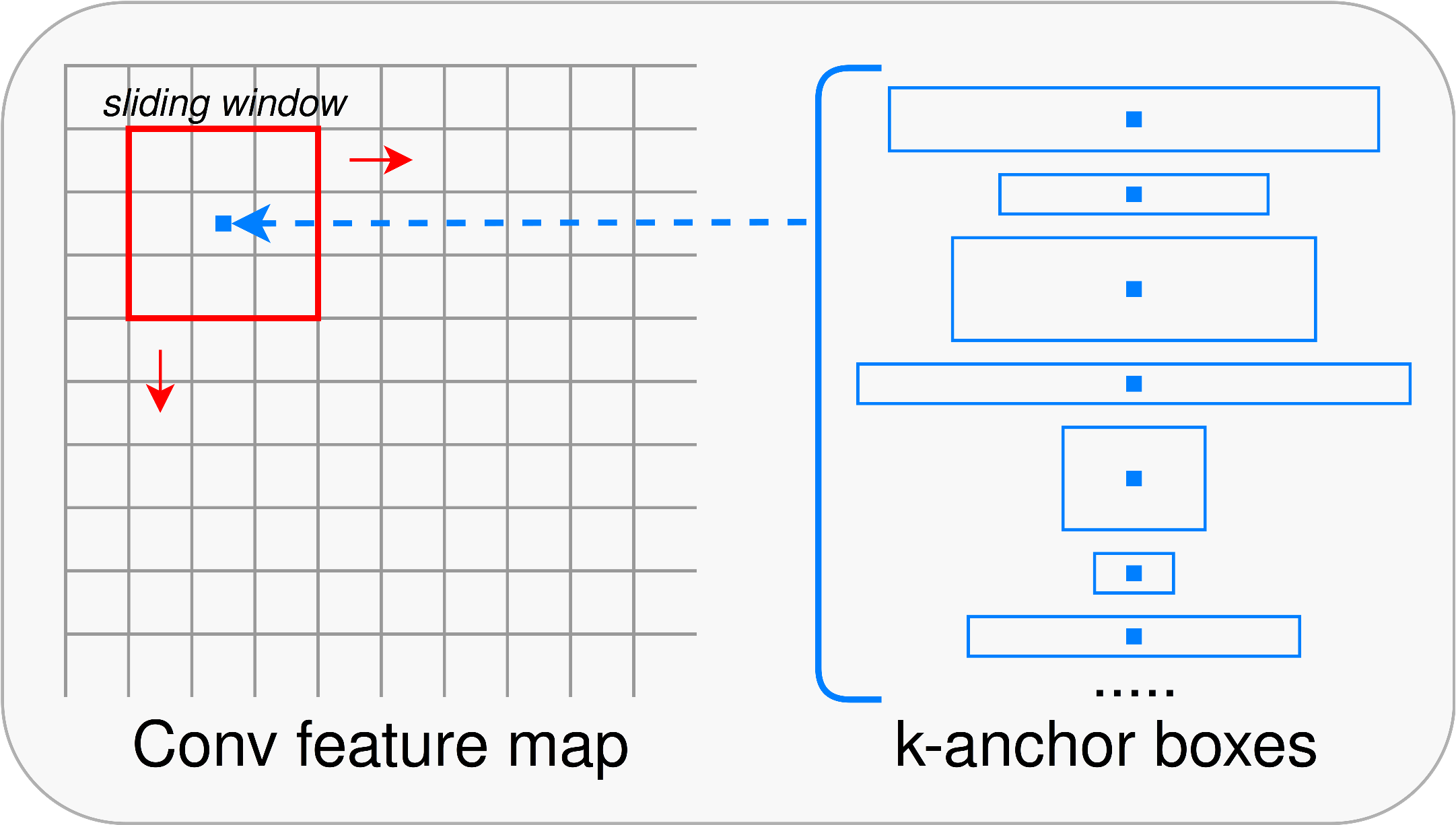}
    \caption[Faster R-CNN region proposal network (RPN).]{A visual representation of the Faster R-CNN region proposal network (RPN): The input image is passed through a deep neural network to extract feature maps, which are then processed by the RPN to generate region proposals. The RPN uses sliding windows and anchor boxes to predict the probability and class of objects in the image \citep{NIPS2015_14bfa6bb}.}
    \label{fig:rpn}
\end{figure}

In R-CNN and fast R-CNN, candidate regions are found by selective search. However, selective search can be time-consuming. Due to this, a new method of improving fast R-CNN has been developed called faster R-CNN. Unlike fast R-CNN, faster R-CNN generates object proposals directly from the image by using a region proposal network (RPN), eliminating the need for selective search or other external proposal generation methods \citep{NIPS2015_14bfa6bb}. An overview of the RPN can be seen in Figure \ref{fig:rpn}.

For each candidate region, faster R-CNN generates a predicted class label and a corresponding bounding box. Thus, the object within this bounding box, or the text block in the case of document layout detection, belongs to the corresponding class label. It is not known, however, which exact pixels within this box are part of the predicted class. This is where Mask R-CNN comes into play \citep{He_2017_ICCV}. Mask R-CNN is built upon the existing stages of faster R-CNN and adds an additional stage that generates a binary mask of the objects. This segmentation mask is predicted using a small fully connected network on a pixel-by-pixel basis. Due to the fact that the prediction is carried out in parallel, this mask branch only adds a small amount of computational overhead. 


\subsection{Tesseract OCR}\label{subsec3}
Tesseract is an optical character recognition engine. It was originally developed by HP between 1984 and 1995 and was made open-source by Google in 2006 \citep{4376991}. Originally a commercial command-line tool, which featured no layout detection and exported unformatted text, it outperformed most other OCR-software in a 1995 test run by UNLV and became a backbone of the Google Books project in the 2000s. Version 3.00 (published in 2010) featured the capacity to export in hOCR, since 2016 neural networks are supported and advanced layout detection (ALTO) has been implemented with the release of Tesseract 4.00. Google withdrew from the further development in 2018, in 2020 archive.org, a large and eminently important stakeholder in the cultural heritage industry, began to use Tesseract instead of AbbyyFineReader, which makes the engine probably the most important and most relevant OCR-tool as of today. It features a wide range of languages and different fonts and allows the development, implementation and integration of particular extensions to tackle specific challenges. 

For Tesseract to be able to detect characters, a couple of steps must be completed. Based on a traditional pipeline, Tesseract’s page layout analysis pipeline starts with connected component analysis. The program organizes blobs into text lines and then breaks lines into words. The actual recognition is performed in a two- stage process. On the first pass, each word is attempted to be recognized, with satisfactory recognitions being passed on as training data to an adaptive classifier. As a result, recognition should improve once a few words have been processed, so an additional pass must be carried out in order to ensure uniform accuracy.

Tesseract detects individual lines using a line-finding algorithm designed to work with skewed pages, thus preserving image quality. To remove drop caps, which often occur in historical documents, and vertically touching characters, a simple percentile height filter is employed in conjunction with blob filtering and line construction. Once the blobs have been filtered, they are assigned to a unique text line, and the baselines are estimated using the least median of squares. The final step is to merge blobs that overlap horizontally. This will result in the correct association of broken characters and diacritical marks with the appropriate baseline.

Word detection requires the separation of two different types of words. Initially, Tesseract tests the detected text lines to determine whether they have a fixed pitch, which means that each character has the exact same width. Tesseract disables additional recognition steps when it discovers such words and chops them into individual characters based on pitch. This is not as straightforward for texts that are not pitched in a fixed manner, since gaps between different words have varying spacings. Tesseract solves this problem by measuring gaps between the baseline and mean line in a limited vertical range. As a result of this stage, spaces that are close to the threshold become fuzzy, requiring the use of additional word recognition steps to arrive at a final decision. These additional steps involve chopping joined characters and re-associating broken characters.

As soon as characters have been broken up into individual elements, each element must be classified in order to be meaningfully utilised. Tesseract’s character classification process involves two steps. During the first step, a class pruner creates a short list of possible class matches based on bit-vectors of classes obtained from a lookup table. Second, each feature calculates the similarity between it and a bit-vector of prototypes of the given class that it may match. During the distance calculation process, the total similarity evidence between features and prototypes is recorded. As a result, the greatest combined distance over all the stored configurations is chosen as the best match for the unknown character.

As of Tesseract version 4.0, released in 2018, several major enhancements and improvements have been made over its predecessors. A significant addition was the introduction of an OCR engine based on long-short-term-memory (LSTM) neural networks, which improved recognition accuracy for complex scripts and languages. At the time of writing, the newest version, version 5.3, is being used. The new version has some minor improvements in recognition over version 4.0, but offers a much better API for custom training. Moreover, it has enhanced its algorithms for analysing text layout, making it more accurate at determining the order and structure of text on a page.

\subsection{Enhancing OCR}\label{sota}
The process of extracting information from historic documents can also be considered as a complete extraction pipeline, instead of individual tasks.
\citep{monnier2020docextractor} present work on an extraction tool for historic documents, which provides improvements in terms of robustness and extraction performance due to mutual reinforcement of text line and image segmentation. 
The task of segmentation is also highlighted in \citep{gruber2020automated}, where the authors also propose to conduct preprocessing of the image before conducting OCR.
Such approaches have already been explored in the past, for example by processing the background of the image~\citep{shen2015improving}.
Consequently, techniques such as Generative Adversarial Networks have been explored to achieve super-resolving of the input images~\citep{lat2018enhancing}. Augmentation has been used on several occasions lately in order to develop economic ways to scale up information extraction from historical documents. \citep{Gruning2019} uses the deep neural network ARU-Net to address the issue of line detection in historical documents. \citep{Martinek2020} realises an approach, which combines fine-tuning OCR-engines with comparatively little data, after training the engine with large synthetic data-sets. 
\\
Document layout analysis has received increasing attention in the past few years, solutions found in this field are not alway tackling specifically historical problems, as for instance \citep{Binmakhashen2019}. It has been recognised though that the increasing availability and usability of deep neural networks, in particular CNNs, offers entirely new opportunities for the development of custom-made solutions for certain document layout analysis tasks \citep{Boillet2020}.

\section{Approach}\label{sec2}
State Manuals and \textit{Hof- und Staatsschematismen} for the Habsburg Empire in particular are commonly provided in PDF format. These documents are accessible via various historical document repositories such as the Austrian National Library \footnote{\url{alex.onb.ac.at}}, the Munich Digitalization Center (MDZ) of the Bavarian State Library\footnote{\url{www.digitale-sammlungen.de}} or archive.org\footnote{\url{www.archive.org}}. Frequently, however, plain text is not available at all or is available in sub-optimal quality. Hence, the objective was the development of a comprehensive approach to significantly increase OCR accuracy in historical documents.  
\\ 
Whereas the general issue of sub-optimal OCR quality is widely acknowledged in historical and further humanist research, most approaches consider layout a secondary line of attack, with very few exceptions, as for instance \citep{Biswas2021} and \citep{Boillet2020}. Even the two probably most important and most widely used out-of-the-box solutions in the field of historical OCR, different distributions of Tesseract and the variety of different OCR and HTR models provided by Transkribus, require significant (manual) effort in data preprocessing, when it comes to extract information from digitised primary sources, which feature structures that can not be detected easily by OCR. Even conventional document layouts already often pose a challenge \citep{Jannidis2017}.  
\\
As we were not only interested in improving the OCR quality, but also in preserving the atomic structure of information entity we encountered in the \textit{Schematismus}, our approach featured a different angle. By interpreting the preservation of the complex original information as context, we further assumed that the provision of this context could contribute crucially to improved OCR accuracy. Generally, particularly in historical texts, OCR is performed page-wise, which means that snippets, belonging to different blocks of text, are performed serially, as common OCR processors work line-wise. 

Thus, we tried to split the individual document pages into their layout elements, in order to preserve the context of the different blocks of text. 
To this end, we used a deep learning convolutional neural network. This is a complex ML algorithm originally developed for object localization and object recognition.
We assumed this approach would be suitable, since it should allow the identification of large coherent chunks of information, grouped into a finite number of recurring blocks. 

In the next step, we used an OCR algorithm to process the individual image snippets, rather than the entire document page image. In addition, we addressed the research question, to which degree OCR accuracy could be improved by fine-tuning the standard OCR tool Tesseract on a custom font, which was designed to look as similar as possible to the original font used in the \textit{Schematismus} documents. 

\begin{figure}[ht]
    \centering
    \includegraphics[width=\columnwidth]{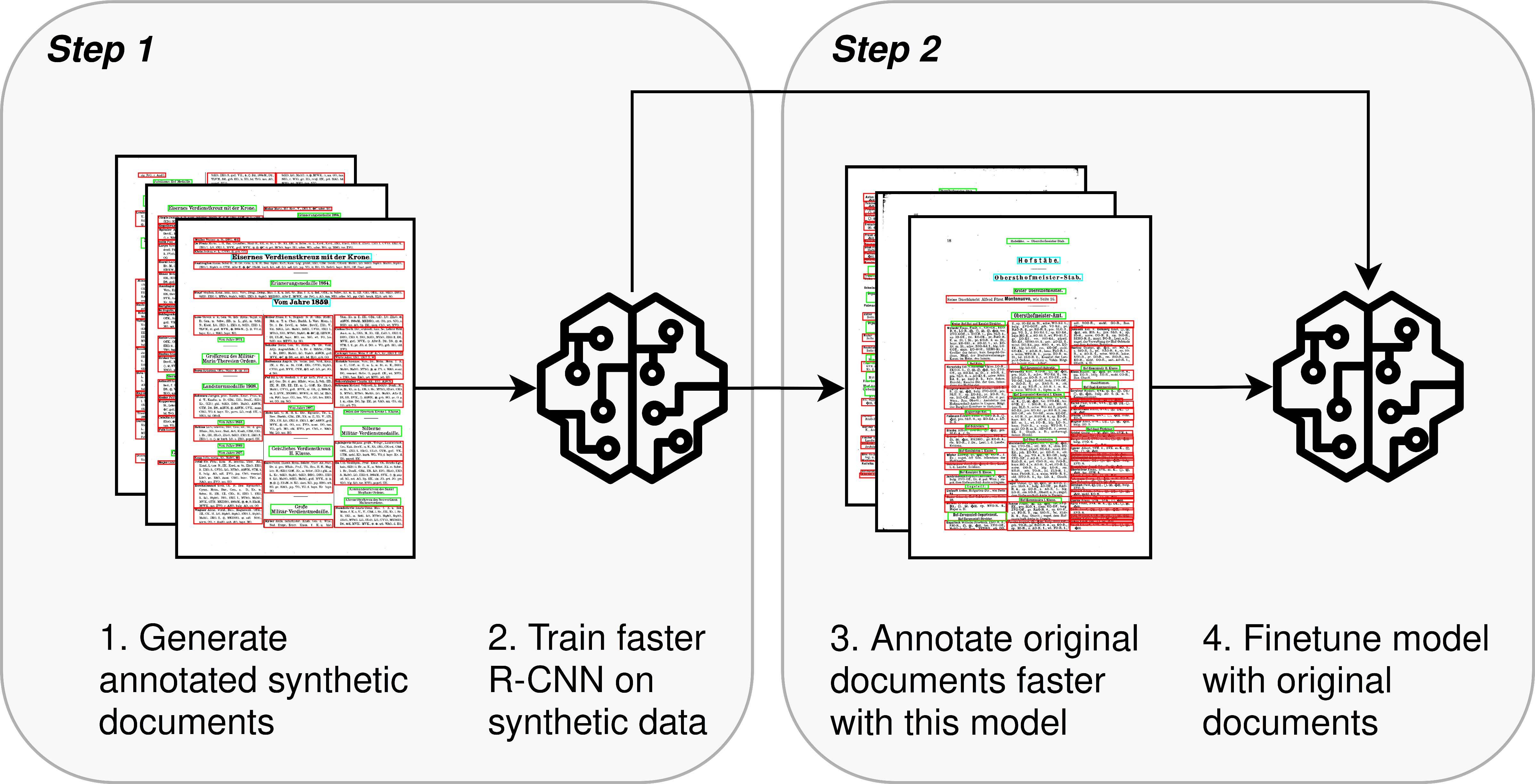}
    \caption[Flowchart showing the general layout detection approach]{The flowchart shows a simplified version of the general process of developing a model to detect layout of schematismus documents.}
    \label{fig:flowchart_layout}
\end{figure}

In a first step, we focused on State Manuals that were published from the second half of the 19\ts{th} century onward, for two reasons: 
\begin{enumerate}
   \item  The task is becoming slightly more complex for the decades prior to 1848, as the fonts that were used are more diverse and complex. We do not consider this a major problem, yet for the proof of concept we were interested in streamlining the entire research process and to eliminate additional complexities that were not in our primary line of attack. 
   \item Even though State Manuals are available for a period of more than 200 years, the mass of data was produced from the 1850s onward, therefore the yield for a solution capable of dealing with documents of this type is expected to be very high. 
\end{enumerate}
In order to obtain the individual layout elements, a layout detection model had to be trained, which requires a large number of annotated documents. 
In general, for supervised machine learning, more available annotated data is expected to yield better results. 
For the purpose of creating these annotations, a two-step approach was chosen, as can be seen in Figure \ref{fig:flowchart_layout}.
\begin{enumerate}
    \item In a first step, fully annotated synthetic documents in the style of the \textit{Hof- und Staatsschematismus} were generated and used to train the deep learning model.
    \item The model, once trained on the synthetic data set, was used for inference in order to faster manually annotate original documents from the \textit{Hof- und Staatsschematismus}. Once a sufficient number of original document pages had been labeled, it was possible to fine-tune the deep learning model.
\end{enumerate} 

\begin{figure}[ht]
    \centering
    \includegraphics[width=\columnwidth]{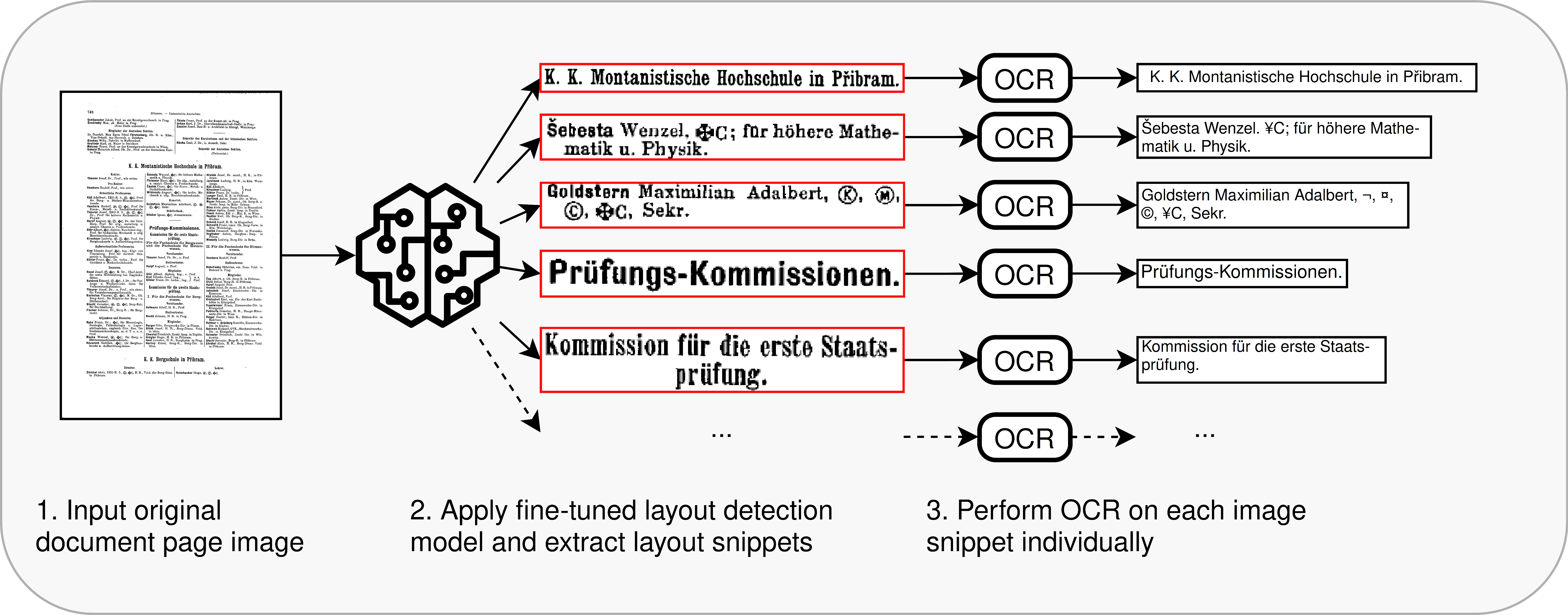}
    \caption[Flowchart showing the general OCR approach]{This flowchart illustrates how each extracted layout element is processed by OCR.}
    \label{fig:flowchart_ocr}
\end{figure}

In the end, we built a machine learning model that can segment retrodigitised PDF-documents of the \textit{Hof- und Staatshandbücher} and split these into their layout-structure elements, such as individual paragraphs and headings. Each of these image snippets was subsequently fed into Tesseract for text extraction. Figure \ref{fig:flowchart_ocr} shows a simplified version of this process.

We used two Tesseract OCR models, one that had been fine-tuned on our custom font, and one that had not been fine-tuned. OCR accuracy was then calculated by comparing the extracted text with the manually transcribed ground truth. Then, this process was repeated, but without dividing the page into individual segments. To answer the research questions in this study, the accuracies were finally compared in order to evaluate the efficiency of our approach.

\section{Implementation}\label{sec:Implementation}
This section discusses in detail how to implement the different methods that we employed to process PDFs of the \textit{Hof- und Staatshandbücher}, including data set generation, layout detection, and optical character recognition. Each of these three steps constituted a work package of its own. All the research and analysis was conducted within Jupyter Notebook, the diverse tools that were put to use are listed in the following subsections. 

\subsection{Data set generation}\label{subsec4}
In order to successfully train a convolutional neural network, a sufficient amount of labeled data is required. Creating a training data set by manually drawing bounding boxes on a large quantity of PDFs drawn from historical source documents is time-consuming and labour-intensive due to the large number of pages that would have to be annotated. Therefore we developed an alternative approach to artificially producing labeled training data. We wrote a Python script that is designed to generate an arbitrary number of synthetic documents mimicking the style of the \textit{Hof- und Staatshandbücher}. This Python script generated Latex-code, which was then compiled using luatex \citep{luatex} to create a PDF file. In the course of this process, the coordinates of the individual text structure elements were determined, which is described in more detail in the following section.

In order for the generated data to be used as training data, it was imperative that the created documents appeared as visually realistic as possible compared to the original documents. Reverse engineering the original documents and paying attention to detail were therefore essential to the creation of synthetic training data.\\

Due to the fact that each paragraph begins with a last name and a first name, a data set containing thousands of first and last names \citep{NameDataset2021} was randomly sampled in order to obtain names. The pool of samples was restricted to Austria, Hungary, Switzerland, and Germany. 

\begin{figure}[ht]
    \centering
    \includegraphics[width=\columnwidth]{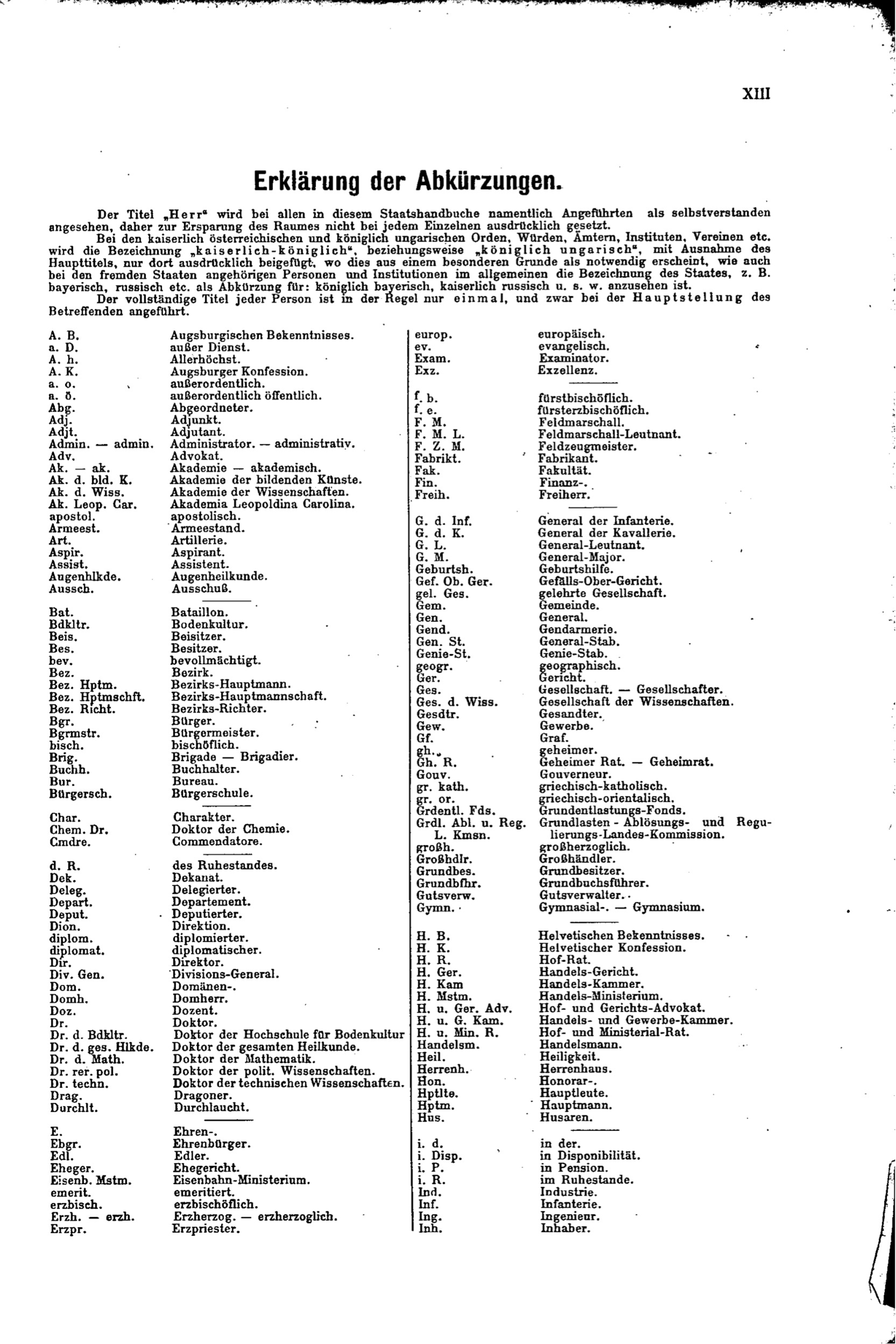}
    \caption[List of abbreviations and explanations.]{A listing of all abbreviations and the corresponding explanation.}
    \label{fig:abbreviations_explanation_dataset_generation}
\end{figure}

A list of abbreviation explanations from the 1910 \textit{Schematismus} document was manually transcribed and randomly selected for the text following the names. The original abbreviation explanation listing can be seen in Figure \ref{fig:abbreviations_explanation_dataset_generation}. To vary the length of individual generated paragraphs, the number of sampled texts was also randomly chosen. 


\begin{figure}[ht]
    \centering
    \includegraphics[width=\columnwidth]{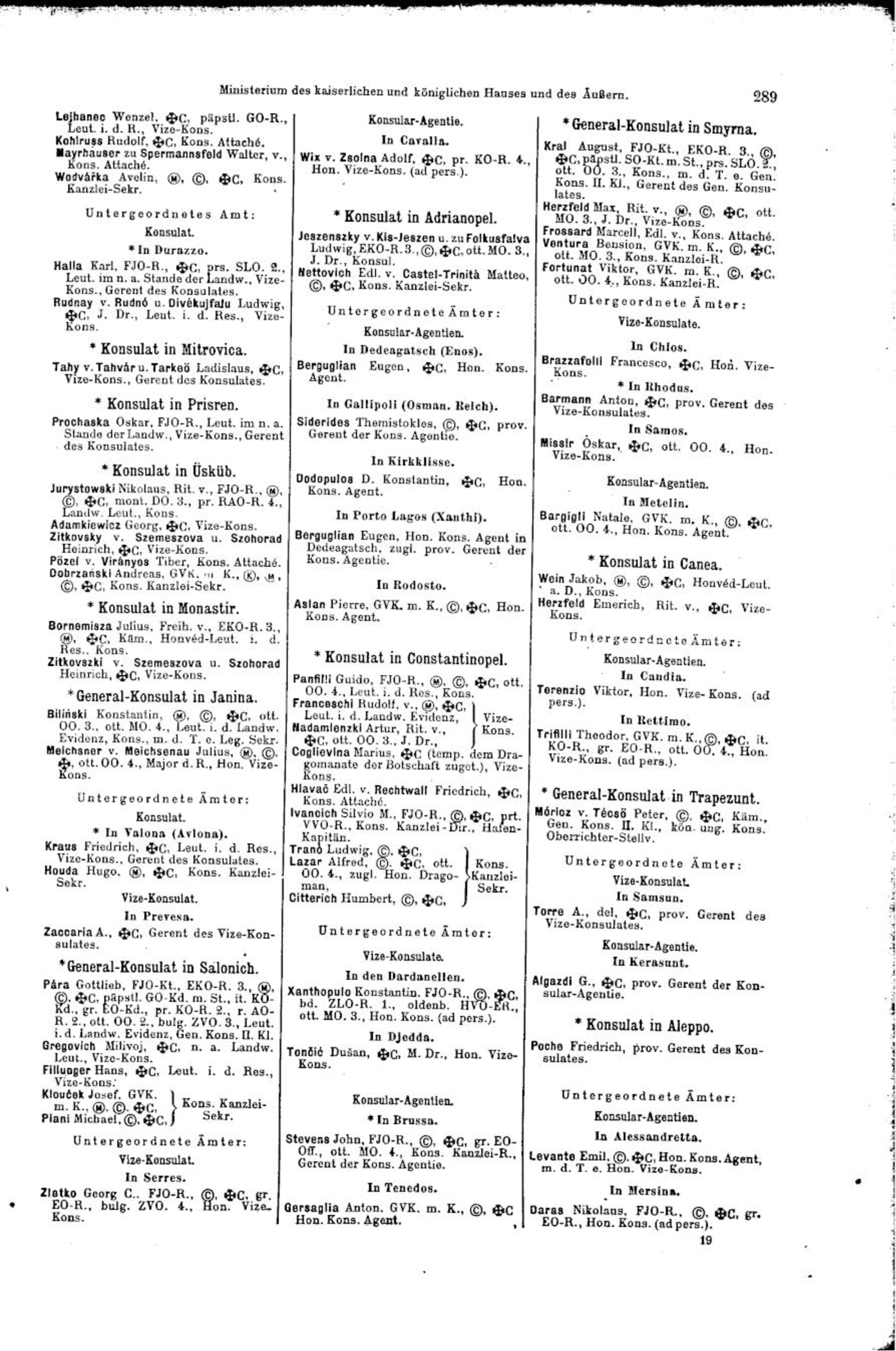}
    \caption[1910 \textit{Schematismus}.]{Example page of \textit{Schematismus} from 1910, highlighting the complexity of its structure, including multiple columns, hierarchical relationships, unique characters, and special special annotations, such as the curly braces of which two of them are found in the middle of the page~\citep{beispiel1910}.}
    \label{fig:example_1910}
\end{figure}

\begin{figure}[ht]
    \centering
    \includegraphics[width=0.4\columnwidth]{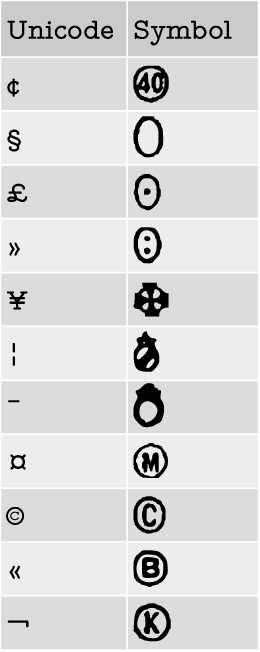}
    \caption[Unicode-mapped symbols.]{Illustration of how Unicode characters are mapped to \textit{Schematismus} symbols.}
    \label{fig:font-mapping}
\end{figure}

\begin{figure}[ht]
        \centering
        \includegraphics[width=\columnwidth]{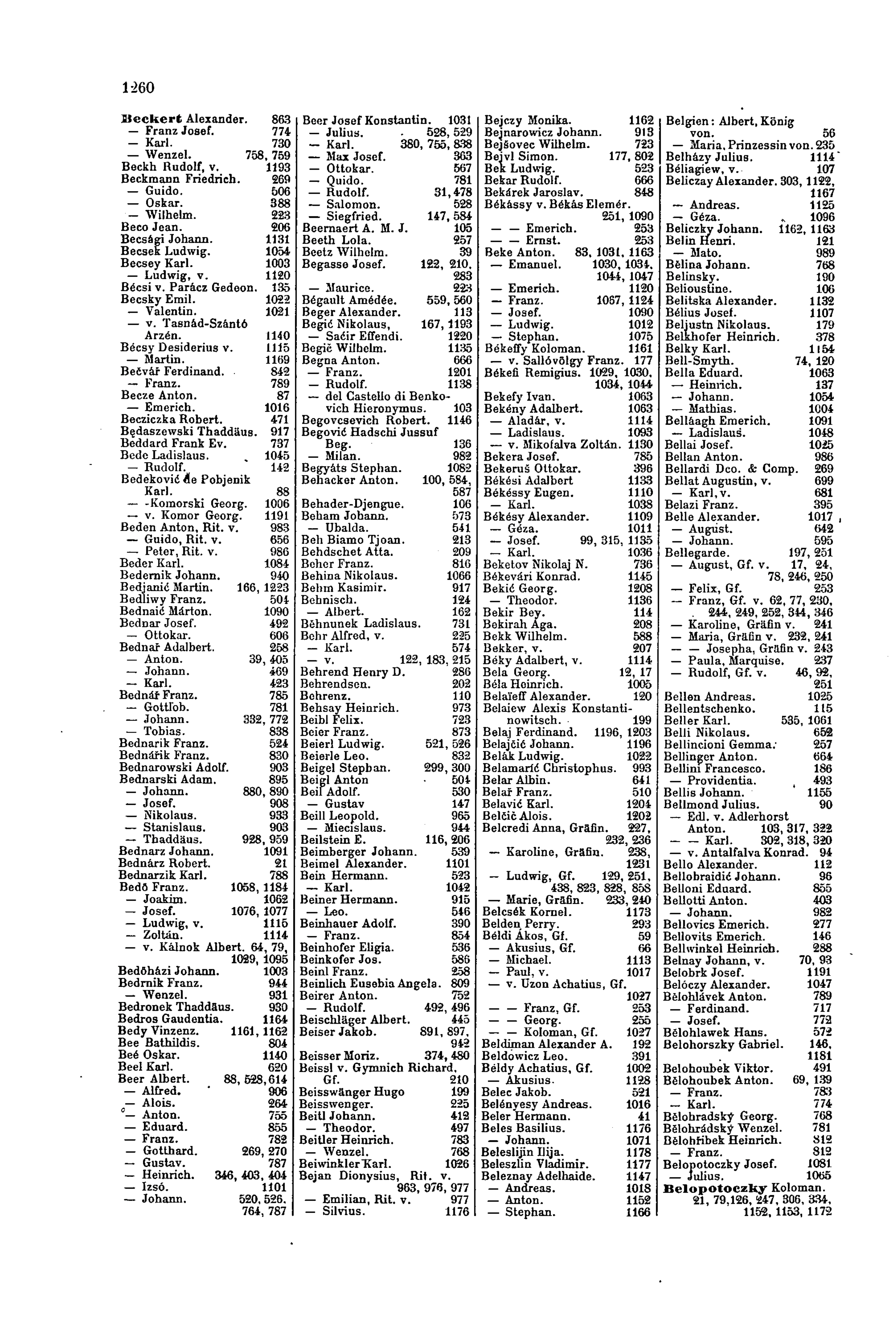}
        \caption[An illustration of an original page from the name index section.]{Example of an original page from the name index section.}
        \label{fig:data set-generation-original-name-index-example}
\end{figure}

Figure \ref{fig:example_1910} displays a page from the original 1910 \textit{Schematismus} document. Variable column numbers are a key characteristic of such documents and are used almost everywhere. While most sections have three columns, there can be variations in certain sections. For instance, Figure \ref{fig:data set-generation-original-name-index-example} shows four columns in the name-index section. 
Thus, generating realistic synthetic documents required the use of the same column layout. \\
Another key visual element is the relatively distinct font type. 
Research led us to a font called ”Opera-Lyrics-Smooth” that appeared very similar to the original. Even though it represented already a good match, we decided to invest additional effort: Using the open source program ”FontForge” \citep{fontforgeFontForge}, we further customised the font in accordance with the original. In order to achieve the best possible match to the original font, screenshots of every letter in the original documents were taken manually, and then the existing letters in the font were adapted according to the screenshots taken. \\
Another distinctive feature of the \textit{Hof- und Staatsschematismus} is its excessive use of particular symbols, representing orders and similar distinctions of the persons listed in this source. The use of these symbols allowed the further condensation of information stored in the \textit{Schematismus}. Nonstandard symbols were mapped to Unicode symbols, which were unlikely to be needed for document generation. Figure \ref{fig:font-mapping} illustrates this mapping.

\begin{figure}[t]
    \centering
        \centering
        \includegraphics[width=0.93\columnwidth]{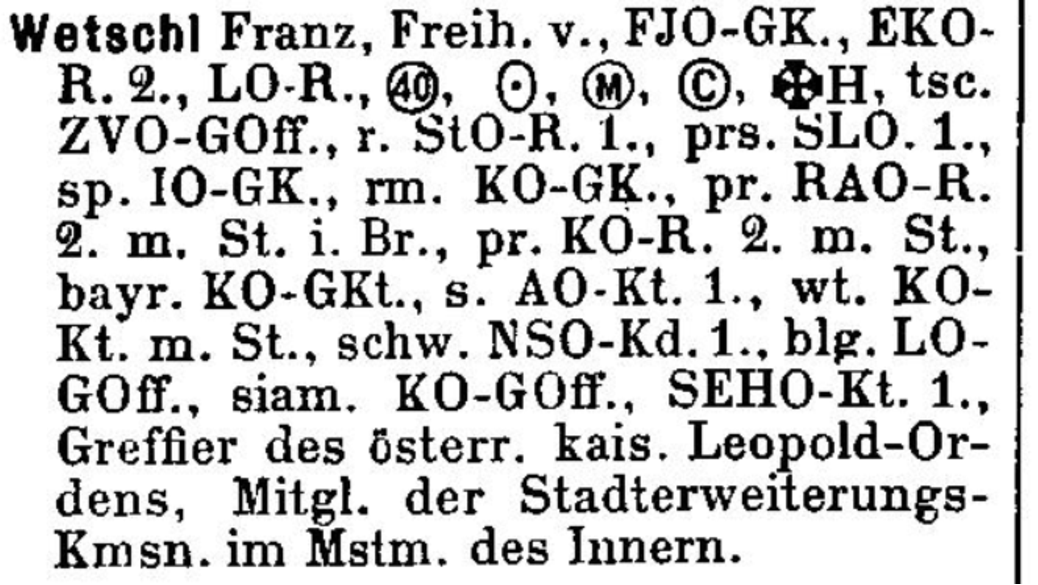}
        \caption{Original paragraph, taken from an existing state manual from the year 1910.}
        \label{fig:example_original_paragraph}
\end{figure}

\begin{figure}[t]
        \centering
        \includegraphics[width=\columnwidth]{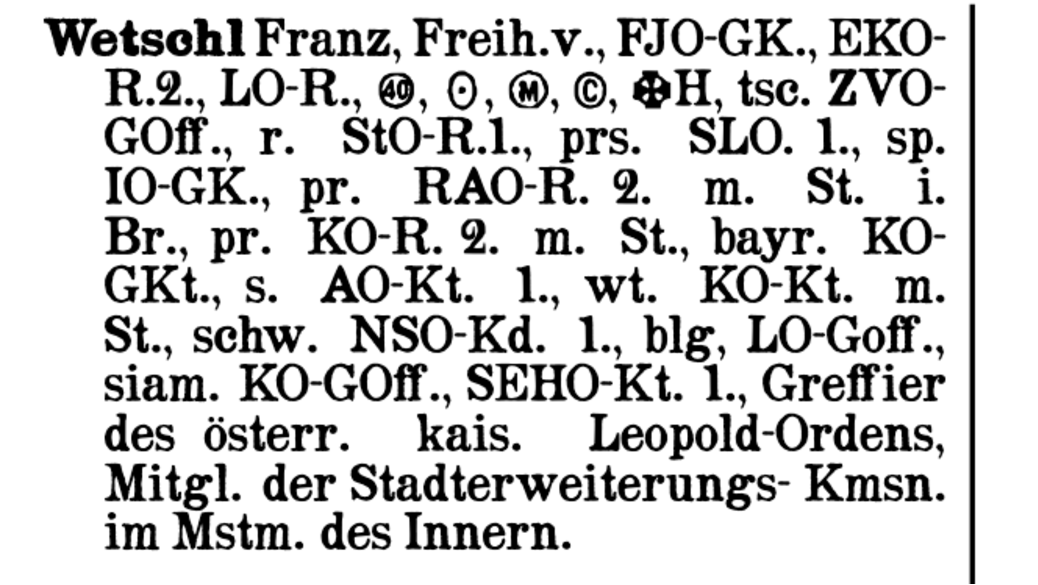}
        \caption{Corresponding synthetically generated paragraph reproduced with a custom font.}
        \label{fig:example_generated_paragraph}
    \label{fig:paragraph-original-generated-comparison}
\end{figure}

This method resulted in the creation of three font types: one for general text paragraphs, one for headlines, and one for italics. An example of this can be seen in Figures \ref{fig:example_original_paragraph} and \ref{fig:example_generated_paragraph}, which shows an original paragraph and an corresponding paragraph reproduced using the custom font.
Additionally, when reviewing some original \textit{Schematismus} documents, it is apparent that font sizes and alignments vary considerably from section to section or even from page to page. In particular, the difference can be observed when examining the headlines of the original documents. To create realistic headlines, four headline types ranging from ”H1” to ”H4” were used to emulate this feature. The first element was the largest and the rest gradually decreased in size. Table \ref{tab:listing_classes} provides a list of all the different classes.\\
Finally, it is crucial to emphasise some small but very significant visual details. Every paragraph begins with one or two words in bold, the last name of the individual, followed by the first name and some additional titles and awards. Indentation occurs after the first line if the text is too long for a single line. Furthermore, multiple individuals may be grouped together within a large curly bracket, as can be seen in Figure \ref{fig:example_1910}. In name-index pages, multiple individuals with the same last name may be grouped by adding a horizontal line at the beginning, which can be observed in Figure \ref{fig:data set-generation-original-name-index-example}. Additionally, every entry within this section is accompanied by one or more numbers that indicate the page number. Finally, it should be noted that headings are usually centered on the page or within columns and that the end of every text is always marked with a period.

\begin{figure}[ht]
        \centering
        \includegraphics[width=\columnwidth]{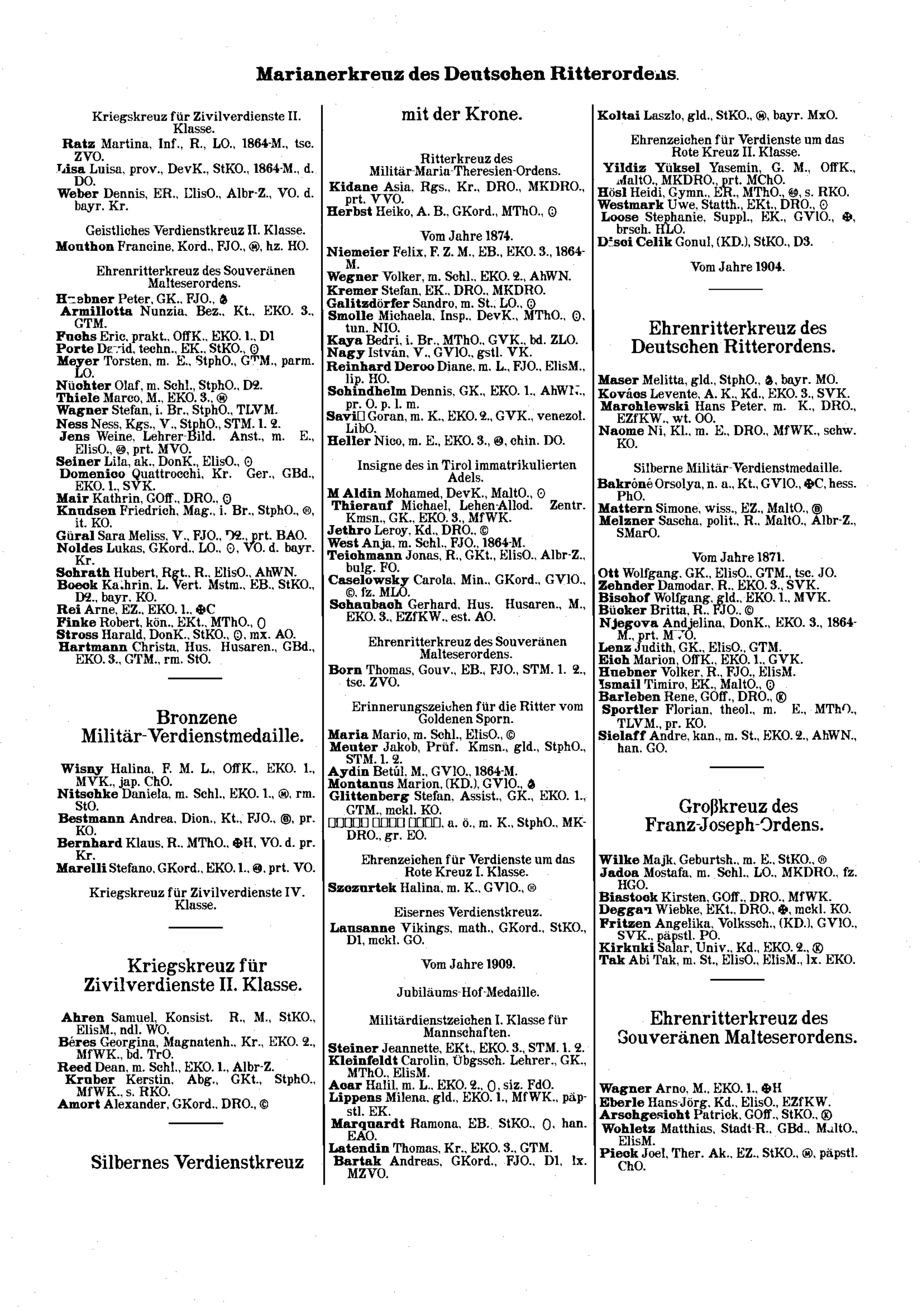}
        \caption[Illustration of a generated \textit{Schematismus}-style document.]{Example of a synthetic \textit{Schematismus}-style document used in training-set.}
        \label{fig:example_generated}
\end{figure}

In order to be able to produce text for the synthetic \textit{Schematismus}-style training documents, several sources were consulted. For the purpose of generating large headlines, a simple list of historical Austrian orders and decorations was used \citep{wiki:Liste_der_österreichischen_Orden_und_Ehrenzeichen}. In order to create headlines with smaller font sizes, a combination of years as strings and Austrian municipality names was used. For paragraph generation, two sources were consulted, as previously mentioned.
Through the use of all the above methods and visual keys, we were able to create a large number of realistic looking synthetic documents. An example of such a synthetic \textit{Schematismus}-style document can be seen in Figure \ref{fig:example_generated}.

In addition to generating a synthetic data set of \textit{Schematismus} documents for training a machine learning model, annotations for each element of the structure had to be generated along with the generation of the document in order to make this data set effective for training.
The additional information consists of bounding boxes with their labels. As part of the process of detecting and localising objects, in this case the layout elements, bounding boxes were used to define the location and size of the individual structure elements within an image. The labels accompanying the bounding boxes indicate the class of the corresponding box, such as ”paragraph” or ”H1”, which are necessary for classification tasks performed by the machine learning method. 

\begin{figure}[ht]
        \centering
        \includegraphics[width=\columnwidth]{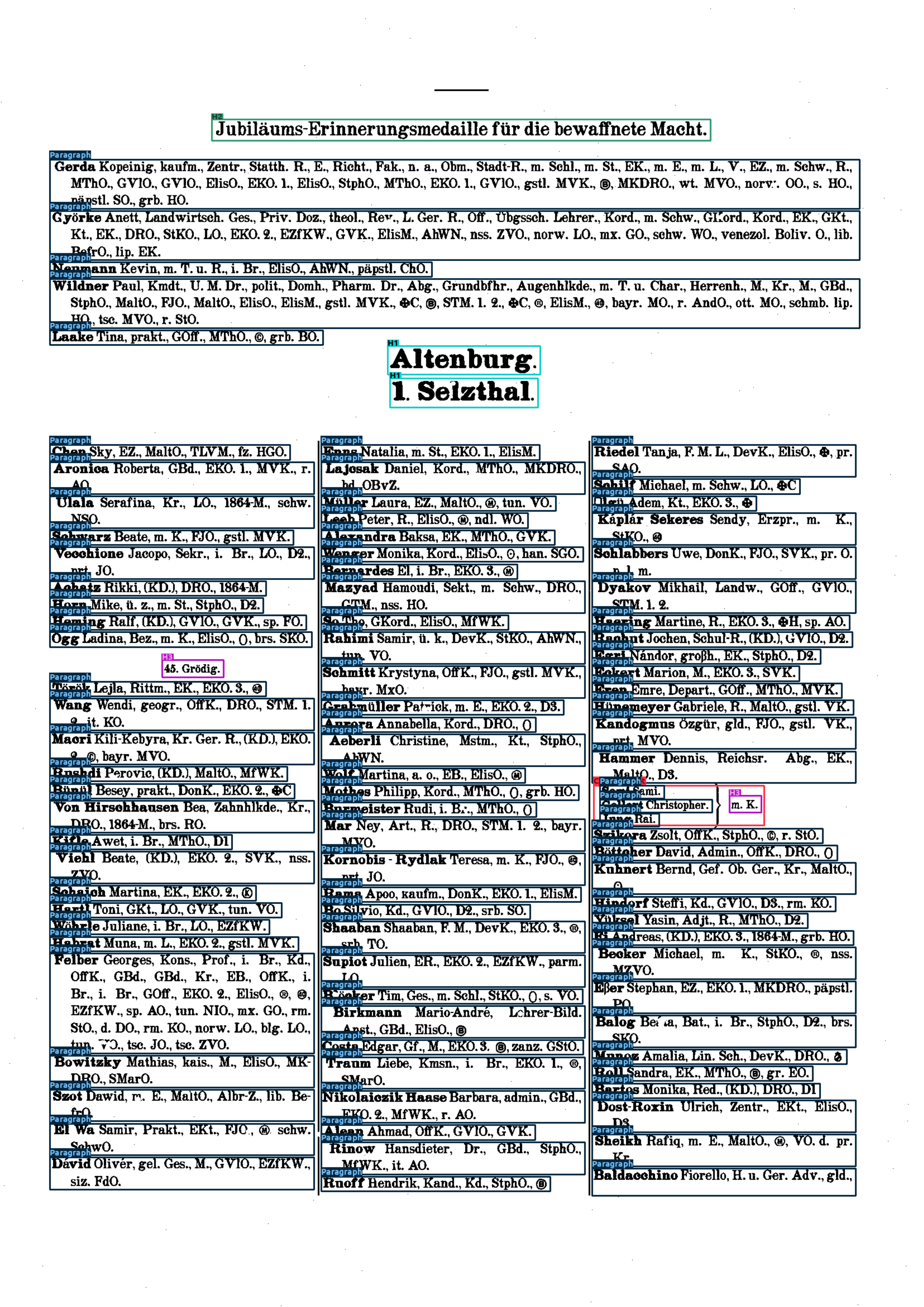}
        \caption[Generated document along with annotations.]{Example of generated \textit{Schematismus}-style document alongside with corresponding annotations (bouding boxes and class label) used in the training set.}
        \label{fig:example_generated_annotated_1}
\end{figure}

A latex package called ”zref-savepos” is used to save the position of characters on the current page and write these coordinates to an external file at compilation time. Using the coordinates, bounding boxes could be computed by parsing the external file. As the individual text elements had already been generated earlier in the same Python script, it was known which label had to be associated with the respective bounding box. In order to construct the data set, the generated documents, which were compiled by the latex compiler and then saved as PDF files, were converted to images and stored in a directory. In addition to the image file, a Pascal VOC XML file containing the corresponding annotations was created and stored in a separate directory. A total of 3,766 synthetic \textit{Schematismus} documents have been generated using this approach. Figure \ref{fig:example_generated_annotated_1} shows a synthetic document with its corresponding annotations overlayed.

\begin{table}[ht]
    \caption{Lists of the class types used to generate synthetic \textit{Schematismus} documents.}
    \label{tab:listing_classes}
    \begin{tabular*}{\columnwidth}{@{\extracolsep\fill}ll}
    \toprule
    \textbf{Class Name} &
      \textbf{Class Description} \\ 
      \midrule
    Paragraph &
      \begin{tabular}[c]{@{}p{5.3cm}@{}}Synthetic documents are primarily composed of paragraphs. Paragraphs always begin with a bolded word and are indented after the first line.\end{tabular} \\
      \addlinespace[0.3cm]
    BigParagraph &
      \begin{tabular}[c]{@{}p{5.3cm}@{}}A big paragraph is a paragraph with a bold starting word but an inverted indentation compared to a normal paragraph. In addition, this type of structure element may never appear within a column.\end{tabular} \\
      \addlinespace[0.3cm]
    H1 &
      \begin{tabular}[c]{@{}p{5.3cm}@{}}H1 headlines represent headings that appear above the start of a column. This class may also never appear within a column.\end{tabular} \\
      \addlinespace[0.3cm]
    H2 &
      \begin{tabular}[c]{@{}p{5.3cm}@{}}H2 headlines have a slightly smaller font size than H1 headings, but will always be contained within a column.\end{tabular} \\
      \addlinespace[0.3cm]
    H3 &
      \begin{tabular}[c]{@{}p{5.3cm}@{}}An H3 header is a title that has the same font size as a paragraph and may only appear within a column. The class is also used to define keywords on the right hand  side of a curly bracket.\end{tabular} \\
      \addlinespace[0.3cm]
    H4 &
      \begin{tabular}[c]{@{}p{5.3cm}@{}}An H4 header is formatted in the same way as an H3, but italicized and always enclosed in parentheses.\end{tabular} \\
      \addlinespace[0.3cm]
    NameEntry &
      \begin{tabular}[c]{@{}p{5.3cm}@{}}Name entries consist of a bold text that is left aligned followed by one or more numbers that are right aligned.\end{tabular} \\
      \addlinespace[0.3cm]
    Curly &
      \begin{tabular}[c]{@{}p{5.3cm}@{}}Using the curly class, multiple paragraphs and a single H3 class can be contained within. In this manner, it should be easier to assign individual paragraphs to a H3 class in the future.\end{tabular} 
      \\ 
      \bottomrule
    \end{tabular*}
\end{table}

\subsection{Layout detection}\label{subsec5}

For the actual layout detection model to be configured in the next step, we chose a faster region-based convolutional neural network (faster R-CNN) built on a ResNet-50 backbone. The model was created and trained using the PyTorch \citep{NEURIPS2019_9015} framework. In PyTorch, version-2 of the faster R-CNN implementation was used \citep{https://doi.org/10.48550/arxiv.2111.11429}. Training of the model was conducted on a Nvidia RTX 4090 with 24GB of video memory.\\
Even though we used primarily the default settings of the model, we found that some adjustments had a significant impact on the model's performance. The following paragraphs will provide a detailed description of these adjustments.
\begin{enumerate}
    \item By setting the pretrained parameter to "True", the training speed has improved in a way that earlier epochs of training have already begun with a lower training and validation loss than with randomly initialised starting weights.
    \item A further parameter that has been tweaked is the anchor generation of the region proposal network in the faster R-CNN model. When identifying areas of interest in an image, anchor boxes are critical because they determine where to look. Thus, they play an essential role when it comes to detecting layout elements within a document image.
    In order to be able to accommodate a variety of different types of objects, anchor boxes were selected with different aspect ratios and scales. It is imperative to note that there are multiple anchor boxes applied to each sliding window position in the region proposal network. Therefore, it is logical to specify these ratios and scales according to the shape of the objects. Thus, the minimum, maximum, and mean ratio and scale of all bounding boxes within the 3,766 generated \textit{Schematismus} documents have been calculated and used as a guideline to set the anchor-generation parameters. A general characteristic of the anchor boxes chosen is their elongated and narrow shape, which is understandable, given that text lines have a similar shape. Additionally, some objects, such as single lines or headings, were quite small, so the anchor boxes generated were smaller than usual.
    \item Further, the maximum number of object detections per image needed to be adjusted. Since most object detection models detect only a few objects at a time within a single image, the default value is 100 objects. It should be noted, however, that since the purpose of this analysis is to detect quite fine-grained layout elements within these \textit{Schematismus} documents, the number of layout elements within one document may easily exceed 250. In order to completely disregard this upper limit of detections per image, the parameter was set to 1000.
    \item Another significant adjustment has been made to the resolution of the images. It should be noted that while the standard image input resolution of the faster R-CNN model is 1,333 by 800 pixels, this resolution results in the model sometimes being unable to detect smaller objects such as single lines within the \textit{Schematismus} documents. The reason for this is that both convolution layers and pooling layers in the model further downscale the input image, resulting in loss of important information. Based on the experiments conducted, a resolution of 1988 x 1405 pixels has been determined as the input resolution.
\end{enumerate}
As mentioned previously, 3,766 synthetic \textit{Schematismus}-style documents have been generated for the purpose of creating a fully annotated data set. Among these, 3,126 served as the training set and 640 served as the validation set. A stratified split of the full data set was used to select the training and validation sets. Stratified splitting involves dividing a data set into two sets while maintaining class distribution in both sets. 
Even though there were more than 3,000 annotated training documents available, dataset augmentation strategies have been applied. 
Adding random augmentations to existing data allowed the training set to be artificially expanded in terms of document variety without increasing the number of documents and thus increasing training time.
Therefore, adding augmentations to the training of a faster R-CNN layout detection model of documents can improve its accuracy and robustness.

As a result of applying random transformations such as rotation, scaling, cropping, optical distortion (to simulate page warping), blur, noise adding and page flipping, the model could learn to recognize and locate different layout elements invariant of their angle or size. Considering that all of these parameters were selected randomly, it is extremely unlikely that two identical documents will be input into the model during training. 
Additionally, augmentation may help to reduce overfitting, which occurs when a model becomes too specialized in recognising only training examples and performs poorly on new, unknown data. By augmenting the training data, the model is exposed to a wider range of layout variations and becomes more adaptable to new and unseen documents.

As for the actual training process, adjustments have been made to the number of epochs, batch size, and learning rate. During each training iteration, batch size determines the number of samples, and thus images, to be processed by the machine learning model before the weights are updated. It is one of the most influential hyper-parameters when training deep learning models, and it can be viewed as a trade-off between accuracy and speed. A larger batch size allows more samples to be processed at once, resulting in faster training times and better hardware utilisation. However, larger batch sizes require more memory and may hinder generalisation \citep{DBLP:journals/corr/KeskarMNST16}.

By contrast, a smaller batch size results in fewer samples being processed at once. Despite slower training times, this can also prevent, to some extent, overfitting and produce a more generalisable model. Typically, smaller batch sizes are used when the data set is small or when the model requires frequent updating of a large number of parameters. Choosing the correct batch size cannot be achieved in a one-size-fits-all manner since it is heavily dependent on the data set being used. For our study, a batch-size of two has been selected based on experimentation since it fully utilises GPU memory and, along with a scaling factor of 85 percent, produces a relatively fast training process.

In order to maintain a constant variance in gradient expectations, it is recommended to multiply the learning rate by $\sqrt{k}$ when multiplying the batch size by $k$ \citep{https://doi.org/10.48550/arxiv.1404.5997}. As a result of extensive learning rate optimisation, a base learning rate of 0.005 has been chosen for batch size one. Thus the final learning rate is $0.005 \cdot \sqrt{2} \approx 0.007$. This learning rate, along with a weight decay of 0.0005 and a momentum of 0.9, was used to initialise a stochastic gradient descent optimiser.
The total number of epochs was set to 100. 
The model is saved after every epoch if the validation loss is less than the previous saved validation loss. In this manner, one can be assured that the final model, which has been trained for 100 epochs, is the one which worked best on the validation set. 

\begin{figure*}[ht]
    \centering
    \subfigure[]{\includegraphics[width=0.42\linewidth]{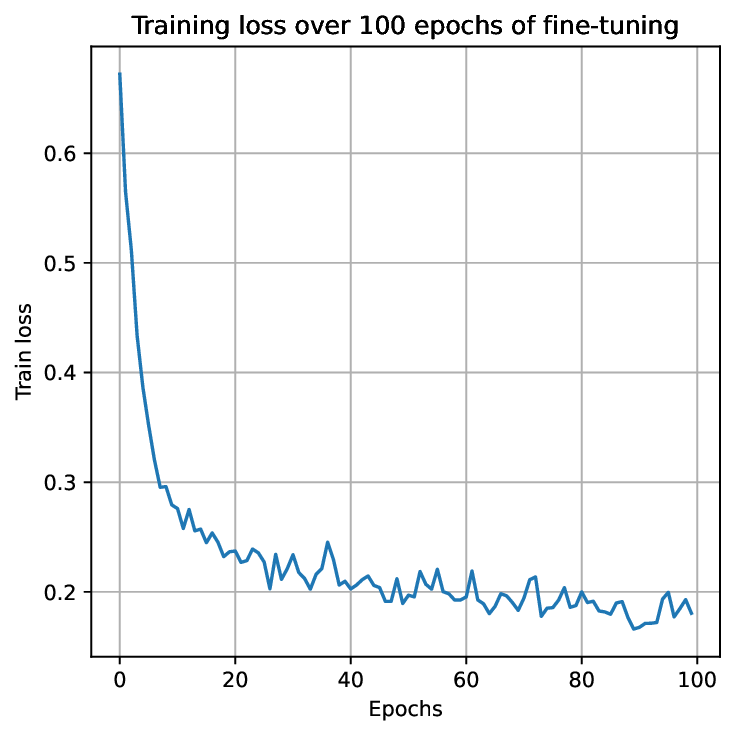}}
    \subfigure[]{\includegraphics[width=0.42\linewidth]{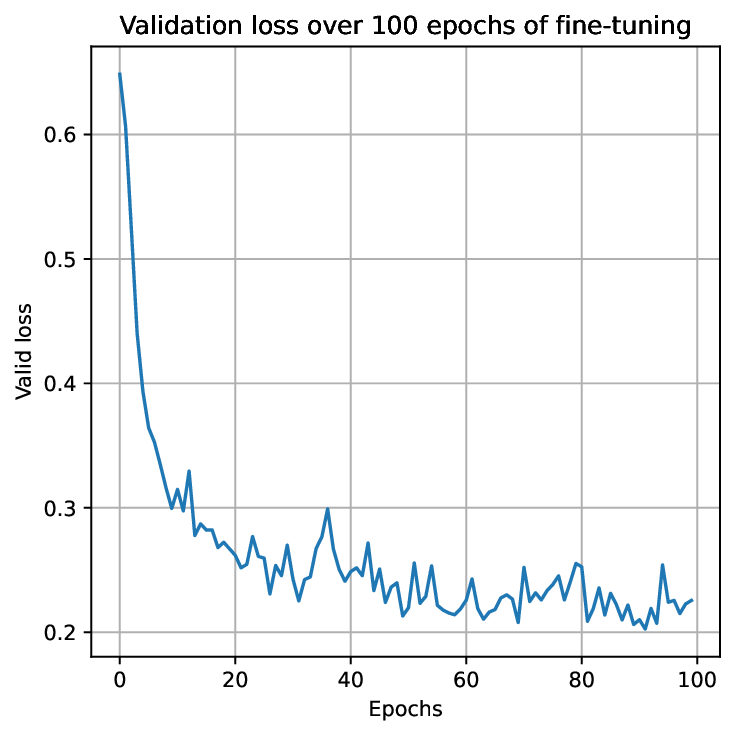}}
    \caption[Training and validation losses for a faster R-CNN model on a dataset of 39 original documents with pre-trained weights.]{This figure illustrates the training and validation losses associated with a faster R-CNN model trained on a dataset containing 39 original documents. Initial weights were derived from the weights of the existing pre-trained model.}
    \label{fig:fine_tuning_losses}
\end{figure*}

The model, which has been trained purely on synthetic data, gave fairly solid results when used on original documents, as described in detail in the \nameref{chap:evaluation} section. 
While these results are already promising, the existing model can also be further fine-tuned using real, original documents. As annotations could not be generated this time, they had to be hand drawn, which is a very time-consuming process. PyTorch's TorchServe \citep{TorchServe} framework has, however, accelerated this process significantly. Employing this method allowed us to "serve" an existing trained model over the local network, where one could send images to and receive bounding-box and label prediction information. In theory, this is not much different from a simple Python script that runs a model directly to predict the layout elements of documents that are fed into it, but there are some very specific applications it can be used for. With the help of an annotation software called "BoundingBoxEditor" \citep{BoundingBoxEditor} the pretrained model can be "served" and therefore accelerate the manual annotation process of the original documents significantly. This is due to the fact that the pretrained model already yielded good results. Therefore, only a few adjustments and error corrections were necessary, such as correcting incorrect classifications or bounding boxes. A total of 39 original \textit{Schematismus} documents have been manually annotated and saved in this way. 
Once a sufficient number
of original \textit{Schematismus} documents had been annotated, the original document images and annotations were utilized to fine-tune the existing model. To that purpose, the newly created data set had been divided into training and validation sets. Using the pretrained model's weights as initial weights, these sets were trained for 100 epochs using the same parameters described earlier. Figure \ref{fig:fine_tuning_losses} show the training loss and validation loss for this training process. 

\subsection{OCR}\label{subsec6}
To extract the text within the individual elements of the predicted layout, Tesseract 5.0 was used \citep{Tesseract}.  As mentioned in Subsection \ref{subsec4}, the font used in \textit{Schematismus} documents is no longer commonly used. Despite the fact that Tesseract comes with a number of pretrained fonts, it makes sense to utilize the custom fonts developed for the generation of \textit{Schematismus} documents to fine-tune the Tesseract optical character recognition model. In addition, due to the symbols that are used in the documents (see Figure \ref{fig:font-mapping}) which are unique to \textit{Schematismus} documents, training on this font is necessary in order to recognise these symbols. In order to fine-tune Tesseract on such a font, images containing a single block of text rendered in the font must be generated. For this, Tesseract's built-in function "text2image" has been utilized. This function requires a large textfile of training text as one of its parameters. Despite the fact that there are existing text files for fine-tuning in German, a custom text file has been compiled using the same text generation method as described in Section \ref{subsec4}. As most of the paragraphs within the original \textit{Schematismus} documents are composed of elements from the abbreviation list shown in Figure \ref{fig:abbreviations_explanation_dataset_generation}, using this information to generate training data for Tesseract is logical. 

In addition, a custom character mapping file called "unicharset" must be provided in order to map the \textit{Schematismus} unique symbols to special Unicode characters. The built-in function has been used to create 50,000 images in total. Additionally to generating individual images containing a single text block and saving them as a ".tif" file, two additional files are generated. One of them is the underlying groundwork, which is saved as a text file. As for the second file, it contains information about every character rendered within the image. It provides details about the character as well as coordinates describing its bounding box. 
The three files are then combined into a single ".lstmf" file, which is essential for the training process once fine-tuning begins. A note should be made regarding the fact that the German OCR model has been used as a starting point for this training process.
Section \ref{chap:evaluation} presents an evaluation of the fine-tuned Tesseract model, as well as additional experiments and preprocessing steps required to obtain the best results.

\section{Evaluation}\label{chap:evaluation}
We start the evaluation of our model performance with an explanation of our parameter choices for the  layout detection model, discussed in subsection \nameref{subsec5}. 
Then, both the Tesseract optical character recognition algorithm and the layout detection model are evaluated in detail. As these two elements perform quite distinctly, evaluation will first take place separately, followed by an evaluation of the combined results in subsection \ref{sec:ocr_layout_detector_evaluation}. Furthermore, any additional pre- or post-processing steps that could be performed to further improve the results will be described. 

\subsection{Evaluation of the layout detection model}\label{sec:layout_detection_evaluation}

\begin{figure*}[ht]
    \centering
    \subfigure[]{\includegraphics[width=0.42\linewidth]{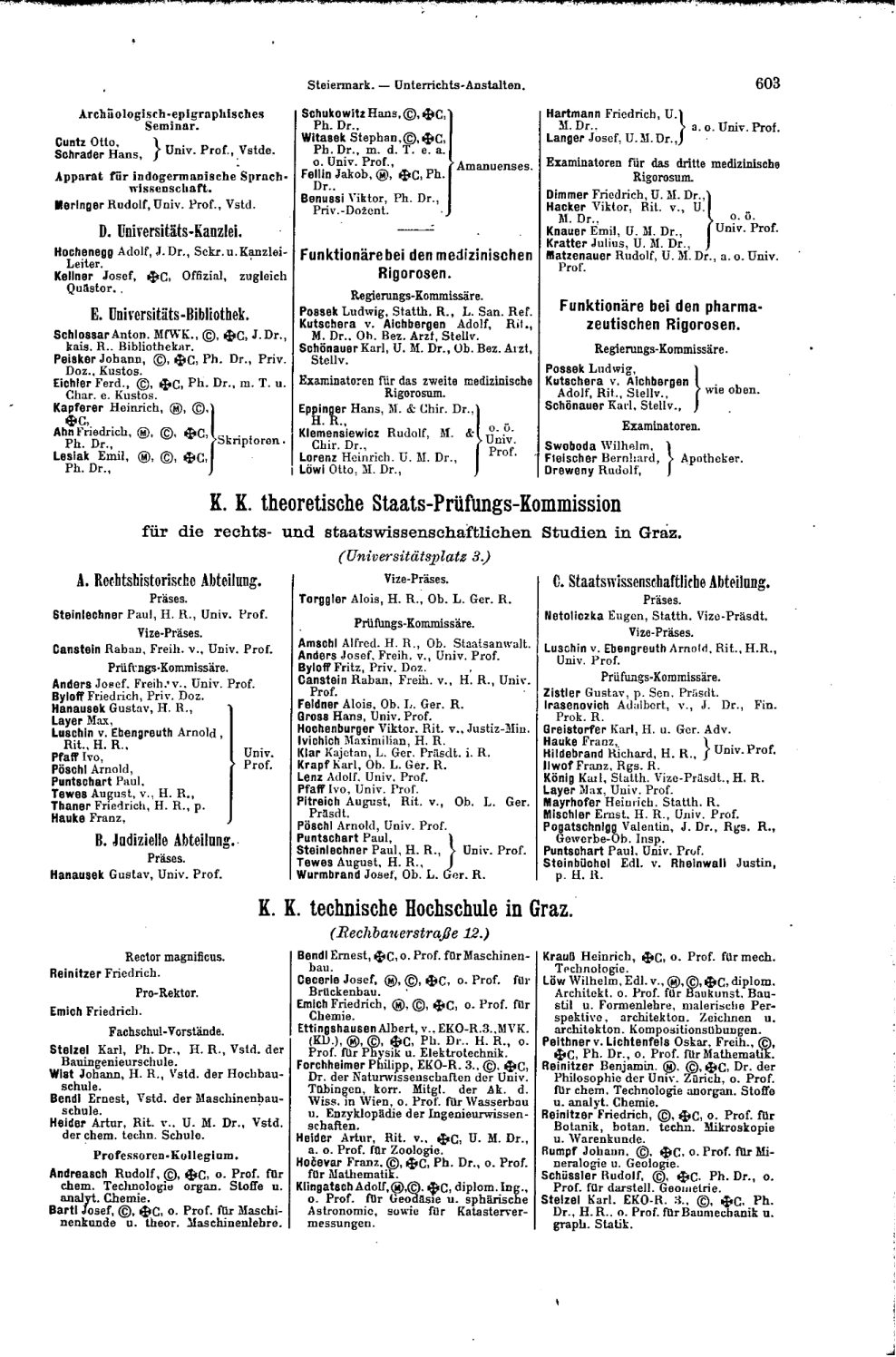}}\quad
    \subfigure[]{\includegraphics[width=0.42\linewidth]{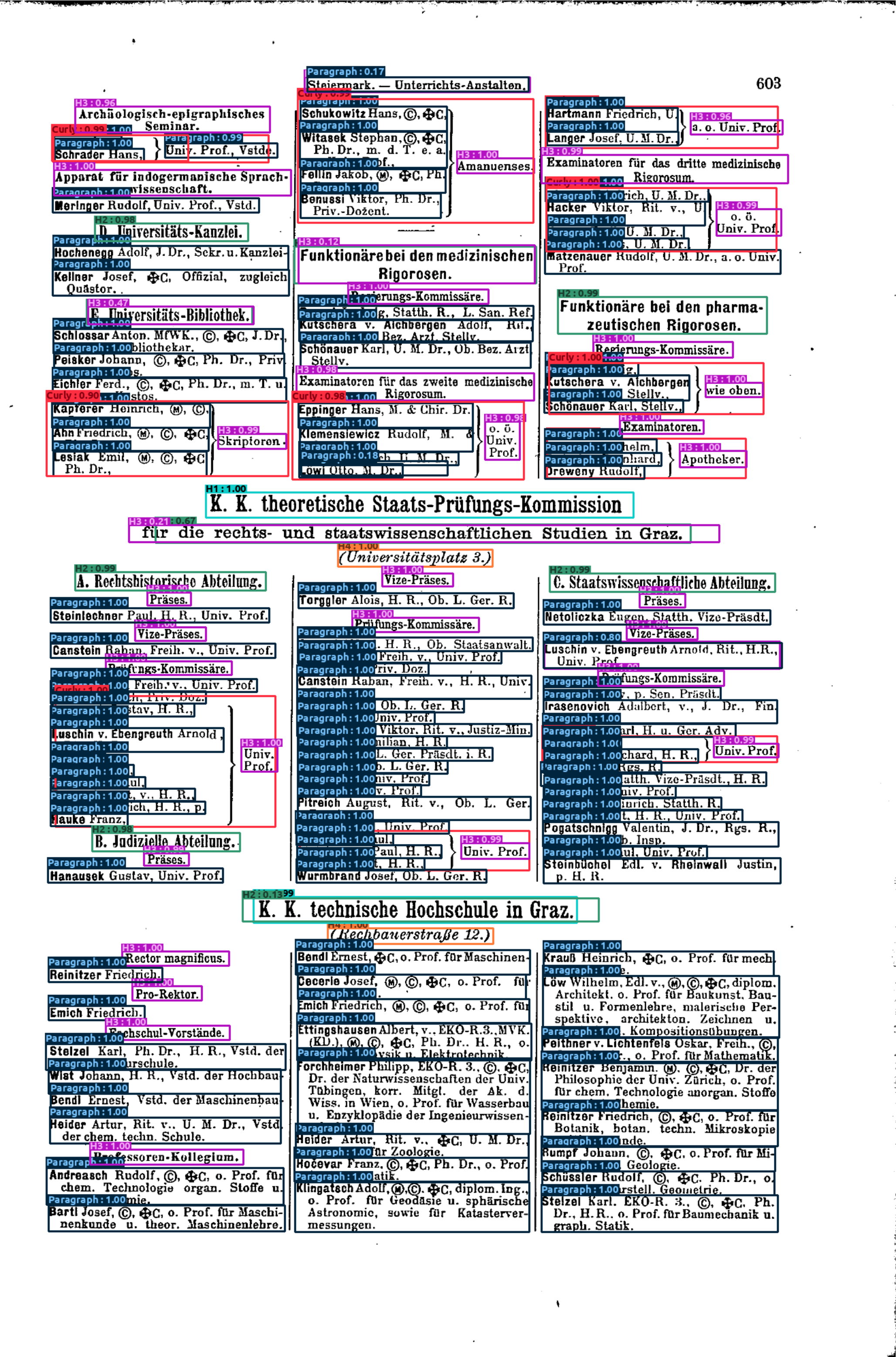}}
    \caption[Original document with and without predicted layout overlayed.]{This figure illustrates a side-by-side view of a randomly selected original \textit{Schematismus} document with and without predicted layout elements overlayed. Element predictions with confidence levels less than 0.1 have been omitted. The faster R-CNN model used to get these prediction results has been fine-tuned on 39 original \textit{Schematismus} documents.}
    \label{fig:evaluation_page_example}
\end{figure*}

\begin{figure}[ht]
    \centering
    \includegraphics[width=1\linewidth]{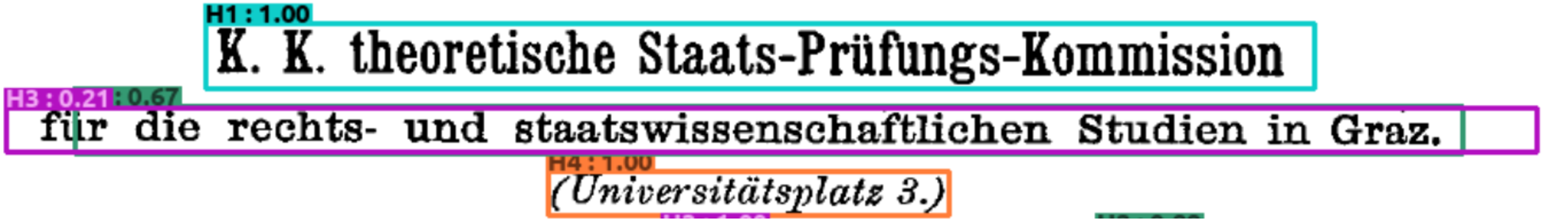}
    \caption[Overlapping bounding boxes with ambiguous labels.]{The figure illustrates that some bounding boxes overlap and have ambiguous label classifications, there are two overlapping bounding boxes in the middel (magenta and teal).}
    \label{fig:model_evaluation_overlapping_boxes}
\end{figure}

A set of fifteen original \textit{Schematismus} document pages from 1910 was analyzed for evaluation of the layout detection model. Eight of these annotated documents were part of the validation set for the fine-tuning process, while the remainder are completely new to the model. It is imperative to point out that these original documents were not part of the training set used to fine-tune the model. An example of how predicted bounding boxes and corresponding labels appear is shown in Figure \ref{fig:evaluation_page_example}. This figure illustrates the model being applied to one of fifteen selected document pages. Predicted elements feature a confidence level between 0 and 1, representing the model's certainty about the accuracy of its prediction. Boxes, which feature a confidence level below 0.1, have been omitted. When analyzing the predictions in detail, it becomes apparent that some bounding boxes are overlapped, leading to sometimes incorrect predictions. An illustration of this phenomenon can be found in Figure \ref{fig:model_evaluation_overlapping_boxes}, in which two different bounding boxes overlap on the second line.

\begin{figure}[ht]
    \centering
    \includegraphics[width=\columnwidth]{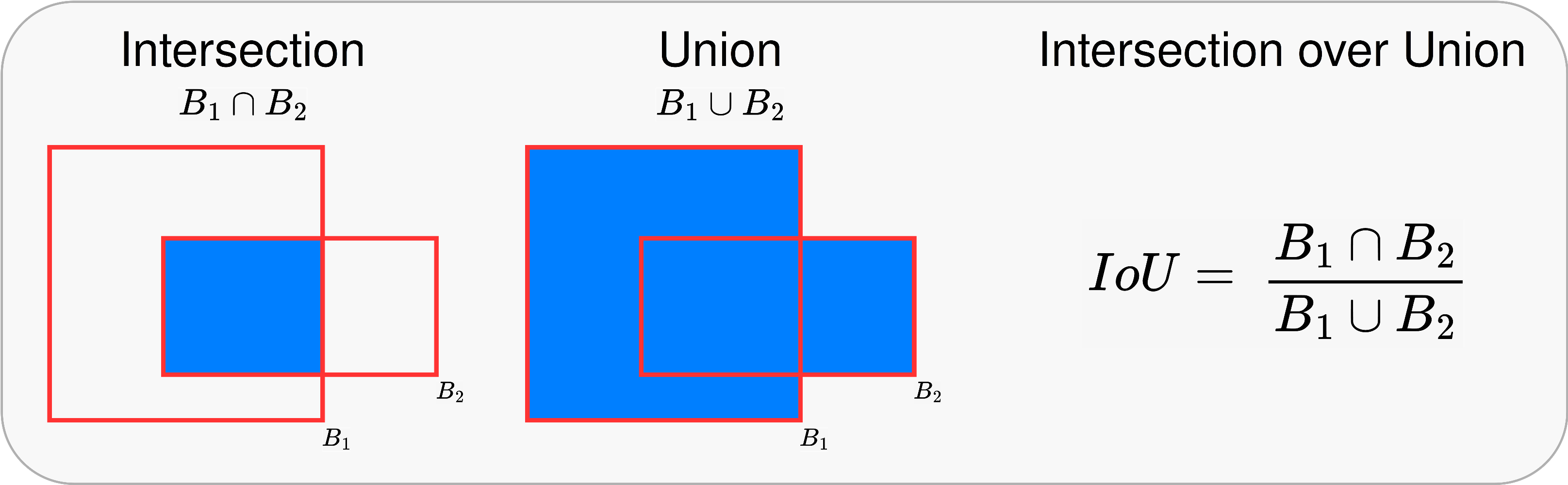}
    \caption[Illustration of the IoU metric used for bounding-box accuracy.]{A visual representation of the intersection over union (IoU) metric is shown in the figure. Furthermore, this metric is used as a bounding-box accuracy metric \citep{zafar2018hands}.}
    \label{fig:IoU}
\end{figure}

To address this issue, a bounding box and label classification post-processing step has been developed. The process works by iterating over every predicted bounding box and then calculating the intersection over union (IoU) with every other bounding box. Essentially, the IoU represents the ratio between the overlapping area and the union area, and the closer the IoU is to 1.0, the more similar the bounding boxes are. A visual representation of this metric can be found in Figure \ref{fig:IoU}. As soon as the IoU has been calculated for every bounding box, merge candidates are identified by selecting boxes with an IoU score higher than 0.3. As a result, a maximum bounding box is calculated, which encompasses all merge candidates (including the box currently being viewed), and is used to replace the original bounding box. Considering that the merge candidates may be of different classes, the merged bounding box will be labelled with the class tag of the highest confidence score. It should be noted that bounding boxes associated with the "Curly" class will not be selected as merge candidates, as the nature of this class is to have enclosed boxes inside them. Figure \ref{fig:model_evaluation_overlapping_boxes} illustrates the overlay of the predicted boxes following the application of the post-processing step.

As a result of this preprocessing step, the layout detection model can now be evaluated. For the purpose of obtaining a ground-truth of bounding-boxes with corresponding labels, the fifteen selected original documents are hand-annotated. During the drawing of the bounding boxes, extensive attention has been paid to detail, in order to reproduce a very accurate ground truth.

\begin{figure}
    \centering
    \includegraphics[width=\columnwidth]{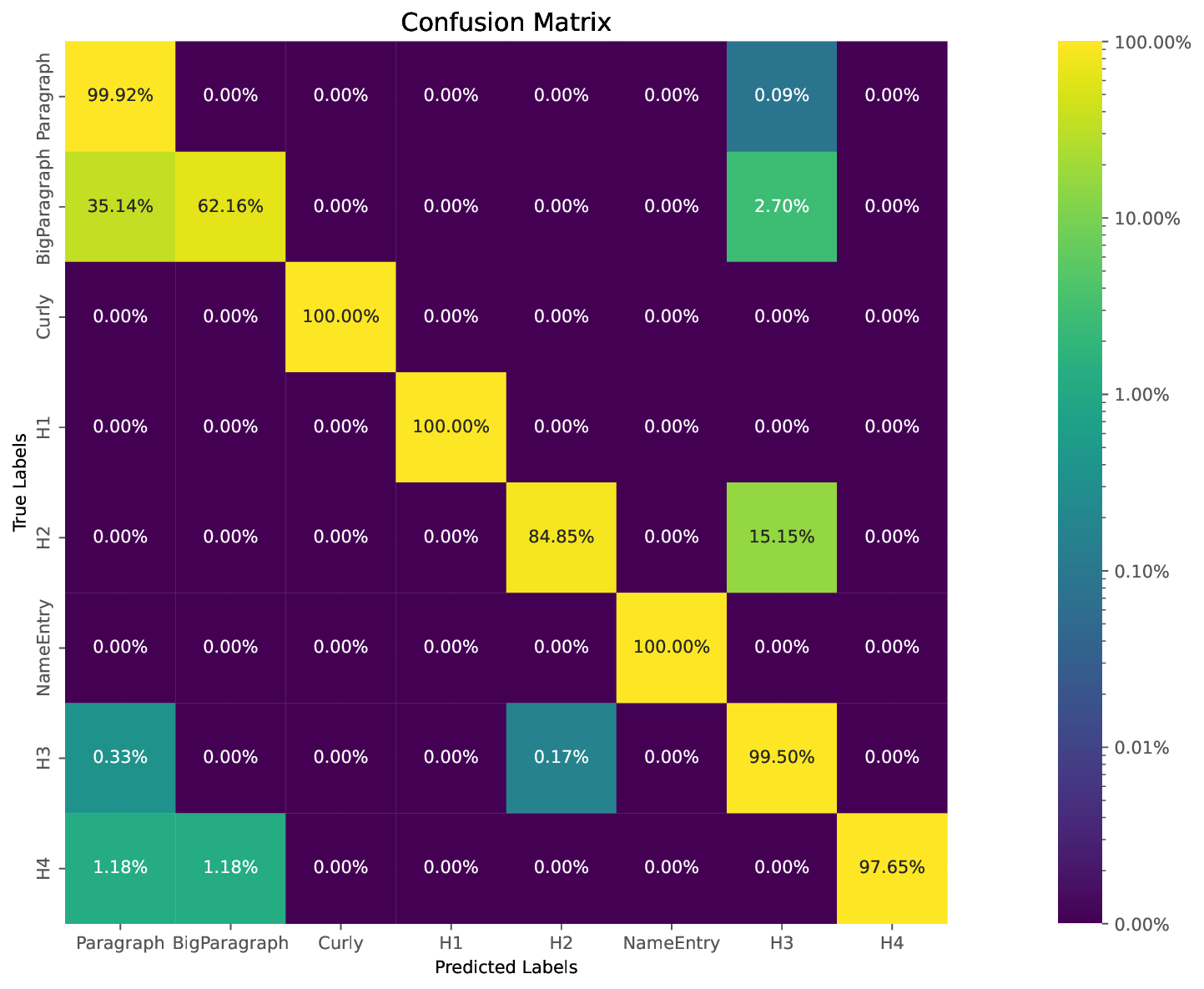}
    \caption[Test set confusion matrix.]{This figure illustrates the confusion matrix for all predictions made on the test set. Note that the values have been normalized according to the ground truth, so that each row sums to 100 percent.}
    \label{fig:confusion_matrix}
\end{figure}

\begin{figure}[ht]
    \centering
    \includegraphics[width=\columnwidth]{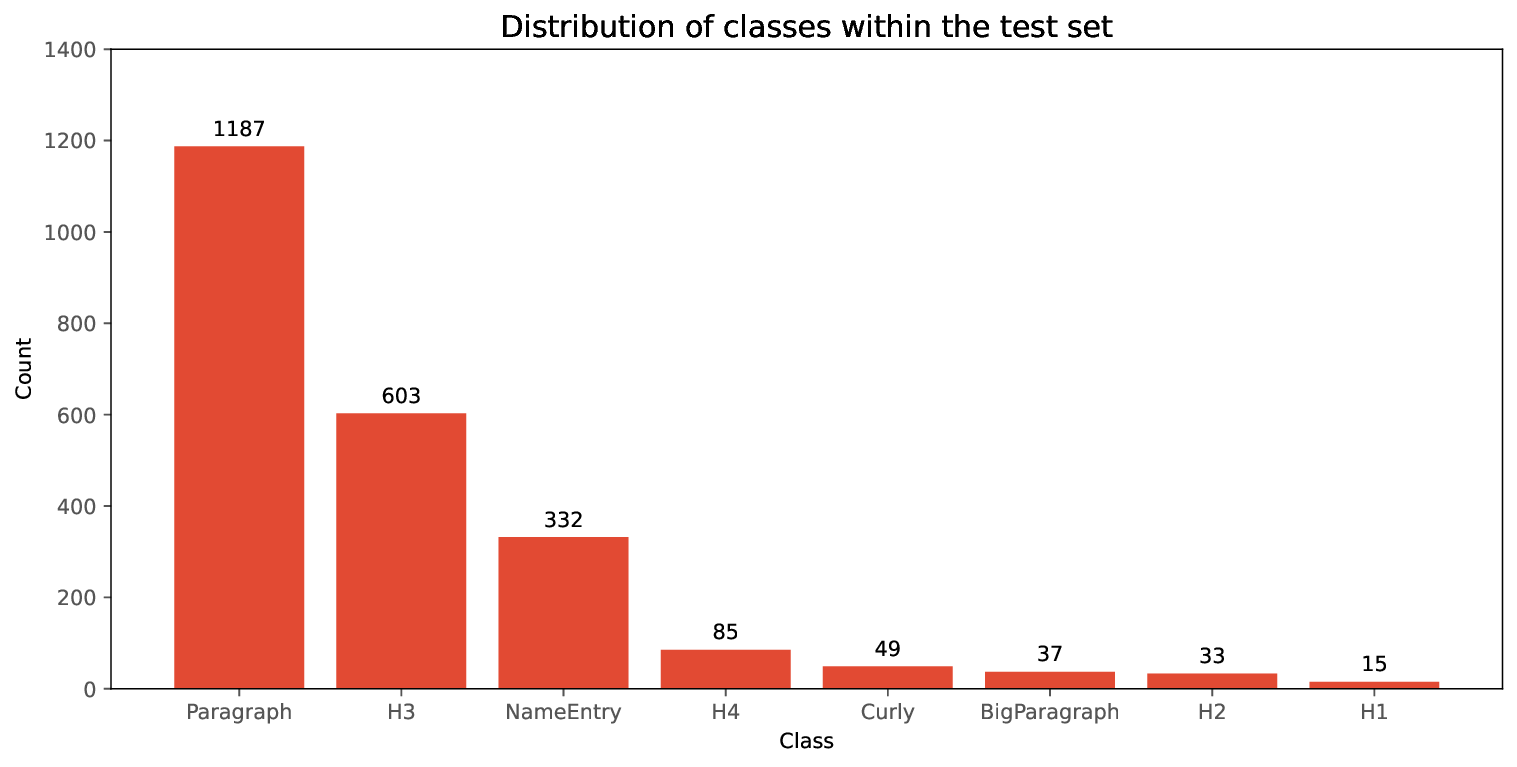}
    \caption[Class distribution of the test set.]{Using ground truth, this figure illustrates the distribution of the various layout elements found in the test set.}
    \label{fig:class_distribution}
\end{figure}

To measure the predicted bounding box accuracy, the intersection over union (IoU) method is again employed (see Figure \ref{fig:IoU}). For each document image in the testing-set, the bounding boxes in the ground-truth set were iterated over and the best matching prediction based on the IoU score was selected. Following the establishment of a list of all best matching predictions, the average over all IoU scores is calculated, which represents the bounding box prediction accuracy for the given page.\\
As these metrics are critical to understanding the following section, true positive, false positive, true negative, and false negative are briefly described in the following. It should be noted that in order to distinguish between binary values in a multiclass classification problem, the metrics have to be calculated for each class individually. This is while only considering the class currently in focus to be positive (1) and all the rest as negative (0). Furthermore, Figure \ref{fig:confusion_matrix} illustrates this with a confusion matrix.
\begin{itemize}
    \item \textbf{TP}\\ A true positive match is one in which both the ground truth and the predicted class are 1.
    \item \textbf{FP}\\ In a false positive match, the actual ground truth is 0 while the predicted value is 1.
    \item \textbf{TN}\\ When both the ground truth and the predicted class are zero, it is considered to be a true negative match.
    \item \textbf{FN}\\ Finally a false negative match is when the ground-truth is 1 and the predicted value is 0.
\end{itemize}
In order to measure the performance of the classification of each layout element, four different metrics are used:
\begin{itemize}
    \item \textbf{Accuracy-Score}\\
    First, there is the classification accuracy score. This is generally considered the most straightforward metric for measuring classification performance. It is essentially a measure of how accurate the model is at categorizing layout elements correctly. For example, if the model categorizes nine out of ten bounding boxes according to ground truth, it has an accuracy score of 0.9 or 90 percent. This concept is illustrated in Equation \ref{eqn:classification_accuracy}.
    \begin{equation}
        \label{eqn:classification_accuracy}
        \text{Accuracy Score} = \frac{TP + TN}{TP + FP + TN + FN}
    \end{equation}

    \item \textbf{Precision-Score}\\
    The classification precision score is the next classification metric used. Prediction precision refers to the ratio of correctly predicted positive observations to the total number of predicted positive observations, thus both true and false positives. In Equation \ref{eqn:classification_precision}, this concept is illustrated. For instance, this metric attempts to determine how many bounding boxes labeled as, for example, paragraphs are actually paragraphs. Although this is trivial for binary classification problems, where only two classes need to be distinguished, it is not as straightforward for multi-class classification problems. Therefore, precision is calculated separately for each class. In order to accomplish this, each class other than the current one is considered negative for each pass. Afterwards, the average precision is calculated across all classes. The resulting metric is referred to as macro-averaged precision. This metric is useful in situations where the performance of a classifier is evaluated in a way that is not biased towards any particular class. When the distribution of classes is highly imbalanced in the dataset, as is the case in this task as a result of the nature of the schematism-state documents (see Figure \ref{fig:class_distribution}), this is particularly relevant, since high precision can be easily achieved for the majority classes while worse performance is expected for the minority classes.
    \begin{equation}
        \label{eqn:classification_precision}
        \text{Precision Score} = \frac{TP}{TP + FP}
    \end{equation}

    \item \textbf{Recall-Score}\\
    The next metric is the classification recall metric. The recall metric, or sensitivity, measures the proportion of correctly predicted positive observations to the total number of anything that should have been predicted as positive. This includes true positives and false negatives, which can be seen in Equation \ref{eqn:classification_recall}. By using this metric, one can determine how many instances of a given class did the classifier correctly identify as belonging to that class out of all of the instances that belong to that class. In the case of a multiclass classification problem, recall is again calculated for each class separately. This is done by considering instances of the same class as positive and all other instances as negative. In order to calculate the overall recall value of the classifier, the recall values of all the classes are averaged.
    \begin{equation}
        \label{eqn:classification_recall}
        \text{Recall Score} = \frac{TP}{TP + FN}
    \end{equation}

    \item \textbf{F1-Score}\\
    F1 is the final classification metric used. The F1 Score is calculated by taking the weighted average of Precision and Recall (harmonic mean), thereby taking into account both false positives and false negatives. The calculation is illustrated in Equation \ref{eqn:classification_f1}. This metric is useful when one is trying to illustrate how precision and recall are balanced in a classification problem. This is because sometimes precision is more relevant than recall, and vice versa.
    \begin{equation}
        \label{eqn:classification_f1}
        \text{F1 Score} = \frac{2 \cdot \text{Precision} \cdot \text{Recall}}{\text{Precision} + \text{Recall}}
    \end{equation}
\end{itemize}

All of the metrics mentioned above have been calculated for each of the 15 documents in the test set in accordance with the ground truth. Table \ref{tab:summary_classification_scores} gives a detailed overview of each metric for both fine-tuned and non-fine-tuned models. Compared to the other documents, documents four and six performed poorly on the fine-tuned model. Explanations and illustrations of why these performed so poorly are provided in Section \ref{sec:error_analysis}. Apart from that, the results for both bounding box accuracy and classification performance appear promising. Table \ref{tab:final_performance_metrics} lists the final performance metrics of the fine-tuned model, averaged across all tested pages.

\begin{table*}[ht]
\caption[Classification accuracy, precision, recall, F1-score, and bounding-box accuracy.]{Summary of classification accuracy, precision, recall, F1-score, and bounding-box accuracy (based on the average IoU) for both non-fine-tuned and fine-tuned models.}
\label{tab:summary_classification_scores}
\setlength{\tabcolsep}{3.5pt}
\begin{tabular*}{\textwidth}{lrrrrrrrrrrrrrrr}
\toprule
\multicolumn{1}{l}{\begin{tabular}{@{}l@{}}\textbf{Document-} \\ \textbf{Index}\end{tabular}} &
  \multicolumn{1}{r}{0} &
  \multicolumn{1}{r}{1} &
  \multicolumn{1}{r}{2} &
  \multicolumn{1}{r}{3} &
  \multicolumn{1}{r}{4} &
  \multicolumn{1}{r}{5} &
  \multicolumn{1}{r}{6} &
  \multicolumn{1}{r}{7} &
  \multicolumn{1}{r}{8} &
  \multicolumn{1}{r}{9} &
  \multicolumn{1}{r}{10} &
  \multicolumn{1}{r}{11} &
  \multicolumn{1}{r}{12} &
  \multicolumn{1}{r}{13} &
  \multicolumn{1}{r}{14} \\
  \midrule
\multicolumn{16}{c}{\textit{FINE-TUNED}} \\
\midrule
\multicolumn{1}{l}{\textbf{Accuracy}} &
  \multicolumn{1}{r}{0.99} &
  \multicolumn{1}{r}{1} &
  \multicolumn{1}{r}{1} &
  \multicolumn{1}{r}{1} &
  \multicolumn{1}{r}{0.74} &
  \multicolumn{1}{r}{1} &
  \multicolumn{1}{r}{0.80} &
  \multicolumn{1}{r}{0.99} &
  \multicolumn{1}{r}{1} &
  \multicolumn{1}{r}{0.99} &
  \multicolumn{1}{r}{0.99} &
  \multicolumn{1}{r}{0.99} &
  \multicolumn{1}{r}{0.99} &
  \multicolumn{1}{r}{0.97} &
  1 \\
\multicolumn{1}{l}{\textbf{Precision}} &
  \multicolumn{1}{r}{0.99} &
  \multicolumn{1}{r}{1} &
  \multicolumn{1}{r}{1} &
  \multicolumn{1}{r}{1} &
  \multicolumn{1}{r}{0.74} &
  \multicolumn{1}{r}{1} &
  \multicolumn{1}{r}{0.56} &
  \multicolumn{1}{r}{0.98} &
  \multicolumn{1}{r}{1} &
  \multicolumn{1}{r}{0.83} &
  \multicolumn{1}{r}{0.99} &
  \multicolumn{1}{r}{0.99} &
  \multicolumn{1}{r}{0.99} &
  \multicolumn{1}{r}{0.82} &
  1 \\
\multicolumn{1}{l}{\textbf{Recall}} &
  \multicolumn{1}{r}{0.98} &
  \multicolumn{1}{r}{1} &
  \multicolumn{1}{r}{1} &
  \multicolumn{1}{r}{1} &
  \multicolumn{1}{r}{0.75} &
  \multicolumn{1}{r}{1} &
  \multicolumn{1}{r}{0.69} &
  \multicolumn{1}{r}{0.88} &
  \multicolumn{1}{r}{1} &
  \multicolumn{1}{r}{0.82} &
  \multicolumn{1}{r}{0.97} &
  \multicolumn{1}{r}{0.99} &
  \multicolumn{1}{r}{0.99} &
  \multicolumn{1}{r}{0.78} &
  1 \\
\multicolumn{1}{l}{\textbf{F1}} &
  \multicolumn{1}{r}{0.99} &
  \multicolumn{1}{r}{1} &
  \multicolumn{1}{r}{1} &
  \multicolumn{1}{r}{1} &
  \multicolumn{1}{r}{0.72} &
  \multicolumn{1}{r}{1} &
  \multicolumn{1}{r}{0.57} &
  \multicolumn{1}{r}{0.91} &
  \multicolumn{1}{r}{1} &
  \multicolumn{1}{r}{0.83} &
  \multicolumn{1}{r}{0.98} &
  \multicolumn{1}{r}{0.99} &
  \multicolumn{1}{r}{0.99} &
  \multicolumn{1}{r}{0.80} &
  1 \\
\multicolumn{1}{l}{\textbf{bbox Accuracy}} &
  \multicolumn{1}{r}{0.91} &
  \multicolumn{1}{r}{0.91} &
  \multicolumn{1}{r}{0.92} &
  \multicolumn{1}{r}{0.92} &
  \multicolumn{1}{r}{0.72} &
  \multicolumn{1}{r}{0.89} &
  \multicolumn{1}{r}{0.88} &
  \multicolumn{1}{r}{0.91} &
  \multicolumn{1}{r}{0.92} &
  \multicolumn{1}{r}{0.93} &
  \multicolumn{1}{r}{0.90} &
  \multicolumn{1}{r}{0.90} &
  \multicolumn{1}{r}{0.88} &
  \multicolumn{1}{r}{0.89} &
  0.87 \\
  \midrule
\multicolumn{16}{c}{\textit{NON FINE-TUNED}} \\ 
\midrule
\multicolumn{1}{l}{\textbf{Accuracy}} &
  \multicolumn{1}{r}{0.80} &
  \multicolumn{1}{r}{0.78} &
  \multicolumn{1}{r}{0.88} &
  \multicolumn{1}{r}{0.65} &
  \multicolumn{1}{r}{0.79} &
  \multicolumn{1}{r}{0.91} &
  \multicolumn{1}{r}{0.50} &
  \multicolumn{1}{r}{0.87} &
  \multicolumn{1}{r}{0.87} &
  \multicolumn{1}{r}{0.75} &
  \multicolumn{1}{r}{0.91} &
  \multicolumn{1}{r}{0.88} &
  \multicolumn{1}{r}{0.10} &
  \multicolumn{1}{r}{0.87} &
  0.96 \\ 
\multicolumn{1}{l}{\textbf{Precision}} &
  \multicolumn{1}{r}{0.51} &
  \multicolumn{1}{r}{0.48} &
  \multicolumn{1}{r}{0.73} &
  \multicolumn{1}{r}{0.36} &
  \multicolumn{1}{r}{0.47} &
  \multicolumn{1}{r}{0.64} &
  \multicolumn{1}{r}{0.50} &
  \multicolumn{1}{r}{0.55} &
  \multicolumn{1}{r}{0.68} &
  \multicolumn{1}{r}{0.77} &
  \multicolumn{1}{r}{0.55} &
  \multicolumn{1}{r}{0.67} &
  \multicolumn{1}{r}{0.49} &
  \multicolumn{1}{r}{0.60} &
  0.50 \\ 
\multicolumn{1}{l}{\textbf{Recall}} &
  \multicolumn{1}{r}{0.53} &
  \multicolumn{1}{r}{0.44} &
  \multicolumn{1}{r}{0.71} &
  \multicolumn{1}{r}{0.32} &
  \multicolumn{1}{r}{0.49} &
  \multicolumn{1}{r}{0.53} &
  \multicolumn{1}{r}{0.38} &
  \multicolumn{1}{r}{0.57} &
  \multicolumn{1}{r}{0.65} &
  \multicolumn{1}{r}{0.77} &
  \multicolumn{1}{r}{0.55} &
  \multicolumn{1}{r}{0.49} &
  \multicolumn{1}{r}{0.06} &
  \multicolumn{1}{r}{0.53} &
  0.48 \\ 
\multicolumn{1}{l}{\textbf{F1}} &
  \multicolumn{1}{r}{0.52} &
  \multicolumn{1}{r}{0.46} &
  \multicolumn{1}{r}{0.70} &
  \multicolumn{1}{r}{0.34} &
  \multicolumn{1}{r}{0.48} &
  \multicolumn{1}{r}{0.56} &
  \multicolumn{1}{r}{0.42} &
  \multicolumn{1}{r}{0.55} &
  \multicolumn{1}{r}{0.66} &
  \multicolumn{1}{r}{0.76} &
  \multicolumn{1}{r}{0.55} &
  \multicolumn{1}{r}{0.55} &
  \multicolumn{1}{r}{0.11} &
  \multicolumn{1}{r}{0.55} &
  0.49 \\
\multicolumn{1}{l}{\textbf{bbox Accuracy}} &
  \multicolumn{1}{r}{0.75} &
  \multicolumn{1}{r}{0.75} &
  \multicolumn{1}{r}{0.76} &
  \multicolumn{1}{r}{0.69} &
  \multicolumn{1}{r}{0.65} &
  \multicolumn{1}{r}{0.71} &
  \multicolumn{1}{r}{0.70} &
  \multicolumn{1}{r}{0.78} &
  \multicolumn{1}{r}{0.79} &
  \multicolumn{1}{r}{0.78} &
  \multicolumn{1}{r}{0.74} &
  \multicolumn{1}{r}{0.68} &
  \multicolumn{1}{r}{0.32} &
  \multicolumn{1}{r}{0.79} &
  0.77 \\ 
  \botrule
\end{tabular*}
\end{table*}

\begin{table}[ht]
    \caption{Performance measures were calculated based on the average of all 15 test documents.}
    \label{tab:final_performance_metrics}
    \begin{tabular*}{\columnwidth}{@{\extracolsep\fill}lr}
    \toprule
    \textbf{Performance Measure} & \textbf{Value} \\
    \midrule
    Average classification accuracy  & 0.964 \\
    Average classification precision & 0.927 \\
    Average classification recall    & 0.923 \\
    Average classification F1        & 0.917 \\
    Average bounding-box accuracy    & 0.889 \\
    \botrule
    \end{tabular*}
\end{table}

Additionally, Table \ref{tab:final_predicton_confidence} provides a statistic about the confidence with which the faster R-CNN model has predicted bounding-box and corresponding class labels for each class. There are three approaches that can be used in order to gain insight into this behavior: The first column in the table represents the average confidence level whenever anything has been detected by the model. Column two indicates the confidence level of the model's prediction, when the associated class was actually correct, and column three indicates the confidence level when the associated class was incorrect. Whenever no incorrect predictions have been made for a particular class, the value has been omitted. Based on the results in column three, it can be seen that the model tends to be overconfident in its predictions.

\begin{table*}[ht]
    \caption{In the Table the confidence levels of the faster R-CNN model in predicting bounding-boxes and corresponding class labels for each class are displayed. Presented are the average confidence level, the confidence level when the associated class was correct, and the confidence level when the associated class was incorrect. According to the results, the model tends to overestimate its predictions, as evidenced by the higher confidence levels associated with incorrect predictions.}
    \label{tab:final_predicton_confidence}
    \begin{tabular*}{\textwidth}{@{\extracolsep\fill}lccc}
        \toprule%
        \textbf{Class Name} & 
        \textbf{\begin{tabular}[c]{@{}c@{}}Average confidence for\\ any prediction\end{tabular}}  & 
        \textbf{\begin{tabular}[c]{@{}c@{}}Average confidence for\\ correct prediction\end{tabular}} & 
        \textbf{\begin{tabular}[c]{@{}c@{}}Average confidence for\\ incorrect prediction\end{tabular}} \\
        \midrule
    Paragraph    & 0.988 & 0.996 & 0.834 \\
    BigParagraph & 0.946 & 0.946 & 0.989 \\
    Curly       & 0.983 & 0.983 & -     \\
    H1          & 0.997 & 0.997 & -     \\
    H2           & 0.884 & 0.918 & 0.647 \\
    H3          & 0.971 & 0.971 & -     \\
    H4          & 0.952 & 0.991 & 0.860 \\
    NameEntry    & 0.988 & 0.988 & -     \\
    \botrule
    \end{tabular*}
\end{table*}

\subsubsection{Experiment: Comparison to non fine-tuned model}
To provide a valuable insight, Table \ref{tab:summary_classification_scores} also compares the calculated metrics with the R-CNN model that has not been fine-tuned on any original documents. On the basis of the results, it can be concluded that fine tuning with a few original \textit{Schematismus} documents, in this case 39, greatly improves the performance.


\subsubsection{Experiment: Layout detection model applied to documents from 1868}

\begin{figure}[ht]
    \centering
    \includegraphics[width=\columnwidth]{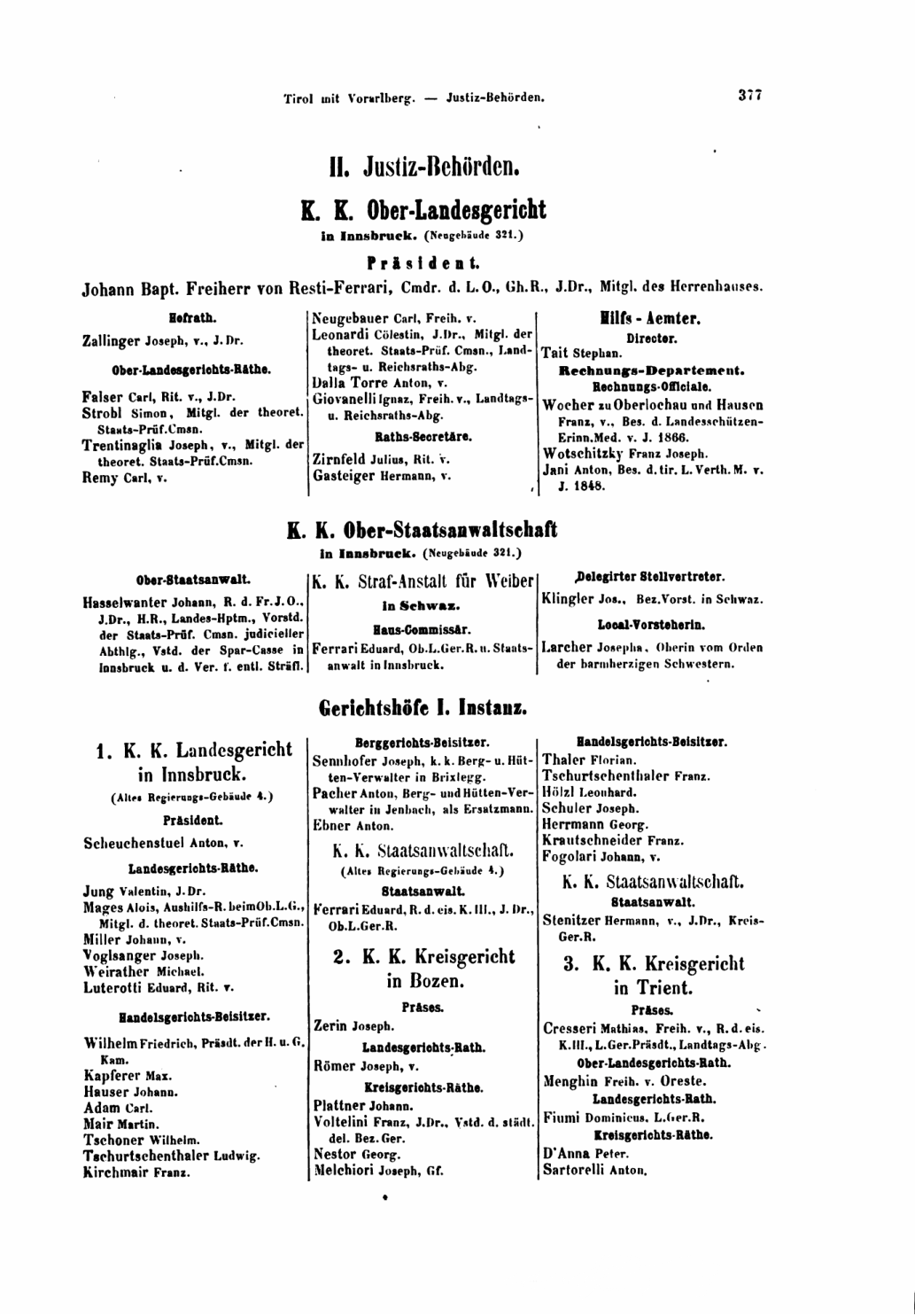}
    \caption{Randomly selected original document from 1868 \textit{Schematismus}.}
    \label{fig:documents_1868}
\end{figure}

To determine how well the fine-tuned layout detection model performs on \textit{Schematismus}-style documents from 1868, two document pages were randomly selected, one of which can be seen in Figure \ref{fig:documents_1868}.
Table \ref{tab:evaluation_metrics_1868} shows the results of the layout-detection model applied to these two documents, and it can be argued through the results that the model performs well with documents from that year. Nevertheless, manual inspection reveals that some paragraphs have been split up by the model, likely due to the different spacing and general layout of these documents compared to those of 1910. 
However, it is expected that with the manual annotation (or generation) of a few documents from 1868 and consecutive fine-tuning these issues can be addressed. 

\begin{table}[ht]
\caption[Results of layout detection for two 1868 documents.]{This table gives the results obtained by applying the fine-tuned layout detection model to two randomly selected \textit{Schematismus} documents from 1868 (see Figure \ref{fig:documents_1868}).}
\label{tab:evaluation_metrics_1868}
\begin{tabular*}{\columnwidth}{@{\extracolsep\fill}lrr}
\toprule
\textbf{Document Index} & \multicolumn{1}{c}{0} & \multicolumn{1}{c}{1} \\ 
\midrule
\textbf{Accuracy}       & 0.743                  & 0.784                  \\
\textbf{Precision}      & 0.741                  & 0.900                    \\
\textbf{Recall}         & 0.903                  & 0.720                   \\
\textbf{F1}             & 0.731                  & 0.746                  \\
\textbf{bbox Accuracy}  & 0.715                  & 0.788                  \\
\botrule
\end{tabular*}
\end{table}

\subsection{Evaluation of OCR performance}

As part of the evaluation of the performance of the Tesseract OCR model, two metrics were used: character error rate (CER) and word error rate (WER). While CER and WER concepts are very similar, CER operates on a character level whereas WER operates on a word level. In order to better understand the meaning of these metrics, it is important to keep in mind that the OCR algorithm can make three types of errors when it extracts text from an image:

\begin{itemize}
    \item \textbf{Insertion error (I)}\\ The output contains additional, and therefore incorrect, characters that are not part of the ground truth.
    \item \textbf{Deletion error (D)}\\ The output contains missing characters, which are present in the ground truth.
    \item \textbf{Substitution error (S)}\\ Characters in the output have been misidentified, resulting in misspellings.
\end{itemize}
In order to determine the character error rate between a prediction and its ground truth, each of the errors listed above must be counted. Therefore, the CER can be calculated using Equation \ref{eqn:cer}, where $N$ represents the number of characters in the reference text. Using this equation, it is possible to calculate the percentage of characters in the reference document that were predicted incorrectly in the OCR output. Thus, a lower CER value indicates better OCR performance at the character level. As a side note, if the predicted output contains more characters than the corresponding ground truth, and thus the numerator in equation \ref{eqn:cer} is larger than the denumerator, this error rate can exceed 100 percent.
\begin{equation}
    \label{eqn:cer}
    CER = \frac{I + D + S}{N}
\end{equation}

According to Equation \ref{eqn:wer}, WER is calculated in a very similar manner. This time, however, substitution, insertion, and deletion errors are not counted at the character level, but at the word level. Therefore, the result represents the number of word substitutions, deletions, or insertions required to transform a sentence into another. As a result, this error rate tends to be higher than CER. This is because it is more stringent, as even a single misspelled character within a word counts as an error.
\begin{equation}
    \label{eqn:wer}
    WER = \frac{I_{W} + D_{W} + S_{W}}{N_{W}}
\end{equation}

To see how Tesseract OCR performs on original \textit{Schematismus} documents, a ground truth must be established. As this ground truth must be compiled manually by converting documents into plain text, which takes a considerable amount of time, only four pages, specifically pages 0, 2, 9 and 13, have been translated from Section \ref{sec:layout_detection_evaluation}.
Then, in a first step, all pages were fed into a distribution of Tesseract OCR, which had not been fine-tuned to the custom font, further, the pages were not preprocessed via layout detection. 
Tesseract was configured to use built-in page segmentation to partition the outputs of the entire pages into an easily readable and correct format. 
The resulting outputs did not match expected sequence. 
Consequently, in order to make a fair comparison between Tesseract's output and the corresponding ground-truth, the predicted outputs have been manually split and reordered to match the original layout of the specific page, without altering any extracted characters or words. This additional process is also responsible for the limited number of pages that have been tested. The exact same process was then repeated using a distribution of Tesseract OCR, which had been fine-tuned for the custom font. Following manual alignment, CER and WER were calculated for every block of text. The average over all pages is shown in rows one and two of Table \ref{tab:cer_wer_final_score}.
According to the results, the CER has improved by 21.62\% and the WER by 19.61\% when using the fine-tuned OCR model. 

\subsection{Evaluation of the OCR in combination with the layout detection model}\label{sec:ocr_layout_detector_evaluation}
In the next step, the layout detection model was utilized to segment every structure element within the pages into individual elements. To obtain the image snippets, the original images were cropped based on the predicted bounding boxes for each page. As each element's coordinates are known in the original document, it was possible to sort them in the appropriate order, so no manual reordering was necessary. The Tesseract OCR model was then applied to each image snippet individually. Since it became evident in the prior step that the fine-tuned version performs significantly better, only this version was used. For each individual image snippet, CER and WER were calculated with the corresponding ground truth. Row three in Table \ref{tab:cer_wer_final_score}, the averages of all predictions across all four pages are presented.

\begin{figure}[ht]
    \centering
        \centering
        \includegraphics[width=\columnwidth]{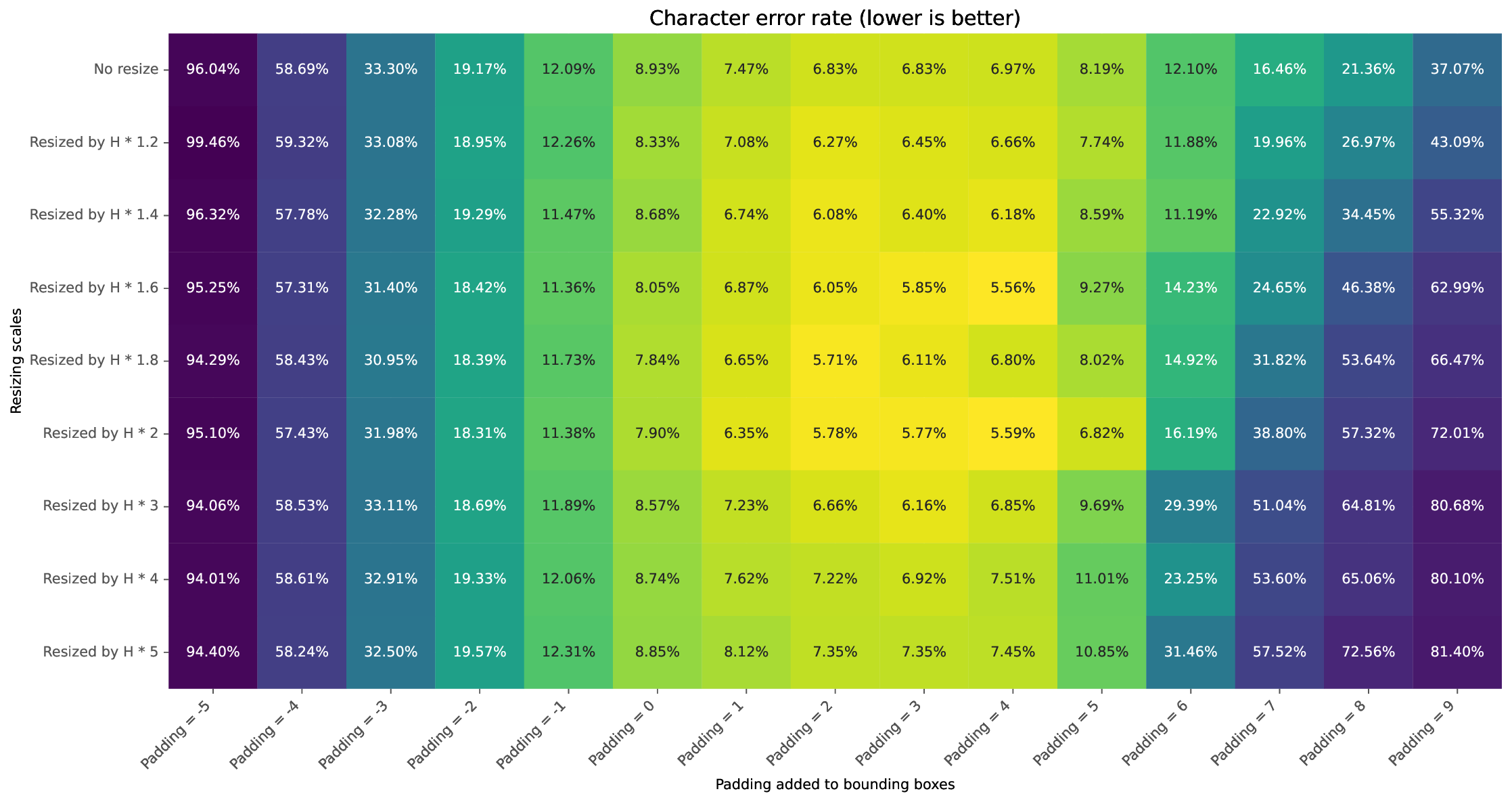}
        \caption{The figure shows the average character-error-rate calculated with different levels of upscaling and padding applied to individual image snippets.}
        \label{fig:cer_matrix}
\end{figure}

\begin{figure}[ht]
        \centering
        \includegraphics[width=\columnwidth]{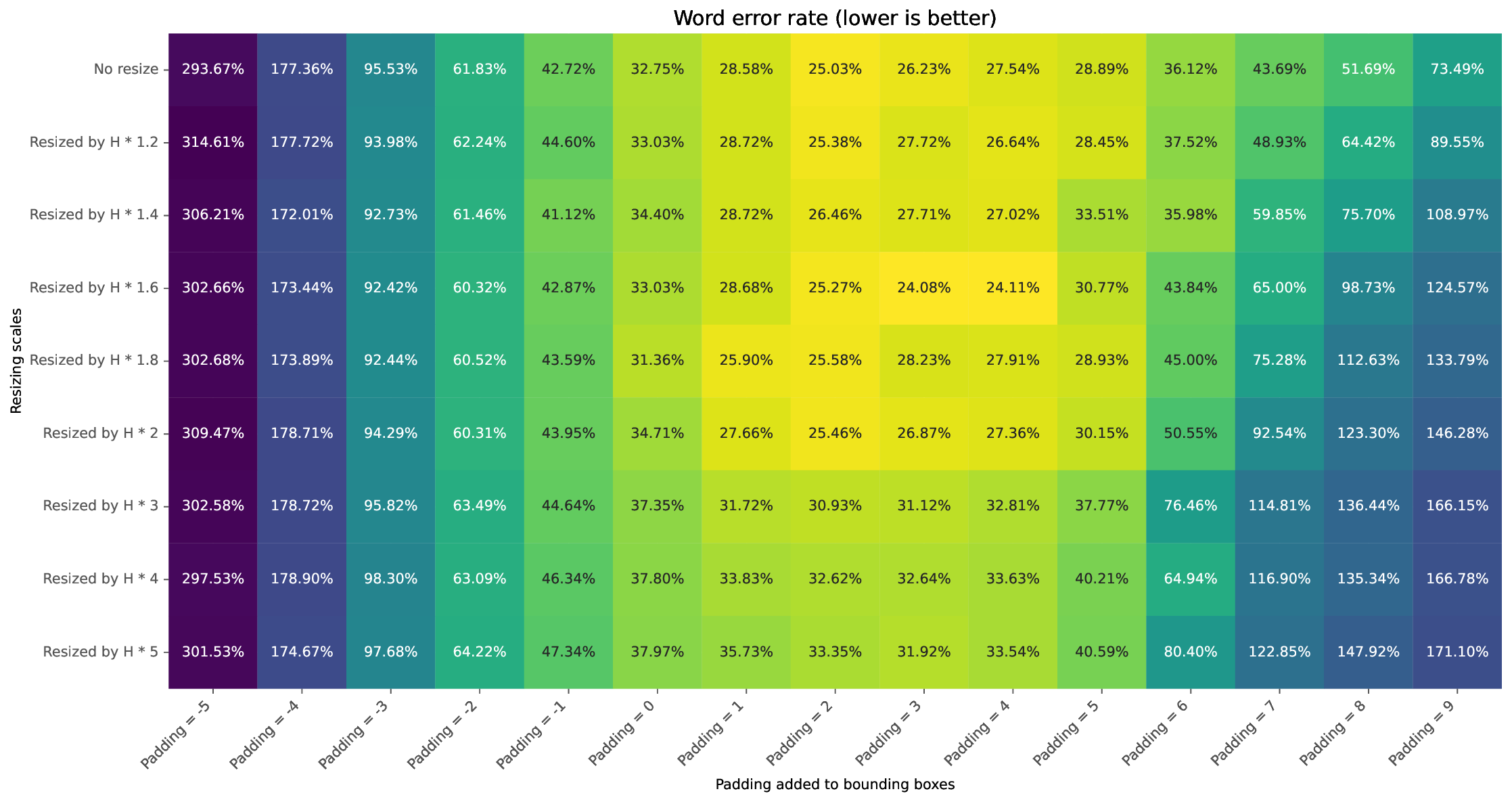}
        \caption{The figure shows the average word-error-rate calculated with different levels of upscaling and padding applied to individual image snippets.}
        \label{fig:wer_matrix}
\end{figure}

According to these results, CER and WER have improved by another 42.57\% and 19.73\% respectively compared to the averages computed based on feeding the full pages into the fine-tuned Tesseract OCR algorithm.
Even though we consider these improvements satisfying, we believe that OCR accuracy can be further improved. 
Although the layout detection model performs well in finding very accurate bounding boxes, sometimes characters are cropped off at the borders of the images. As a result, padding has been added around the predicted bounding box, to address this issue. As described in the official Tesseract guide on improving OCR accuracy \citep{TesseractImproveOCR}, Tesseract generally works better with higher resolution images. Therefore, the individual image snippets have been resized using various scales based on the height of each image in order to find the sweet spot. The aspect ratio of the original image was maintained while upscaling through linear interpolation. Figure \ref{fig:cer_matrix} and Figure \ref{fig:wer_matrix} illustrate this. Interestingly, the CER is at its lowest at a scaling factor of 1.6 times the original image snippet's height with a padding of four pixels. In contrast, the WER is lowest when the same upsampling scale is applied with three-pixel padding. In order to settle for one setting, a padding of four pixels and a resizing scale of 1.6 have been chosen. Row four in Table \ref{tab:cer_wer_final_score} shows the final average CER and WER values.

According to these results, it is possible to answer the research question, which is how much OCR accuracy can be improved by using a layout detection model as a preprocessing step to segment schematism-state documents, and feeding Tesseract individual images containing one layout structure rather than a full page.
In comparison to the average CER and WER obtained on a full page using a fine-tuned Tesseract OCR model, a 64.24\% improvement in the CER and 40.91\% improvement in the WER were observed. In comparison with an out of the box Tesseract OCR model, even higher CER and WER improvements were achieved, 71.98\% and 52.49\%, respectively.

\begin{table}[ht]
\caption{The table presents the average CER and WER resulting from various scenarios. The first two rows show the impact of fine-tuning Tesseract to a custom font on CER and WER on a full page. The third row presents the results when layout detection is combined with a fine-tuned Tesseract OCR model. Finally, in the last row, the results after a padding of four pixels and a resizing scale of 1.6 are applied to each snipped image, are shown.}
\label{tab:cer_wer_final_score}
\begin{tabular*}{\columnwidth}{@{\extracolsep\fill}lrr}
\toprule
& \multicolumn{1}{c}{\textbf{Avg. CER}} & \multicolumn{1}{c}{\textbf{Avg. WER}} \\
\midrule
\multicolumn{1}{l}{\textbf{\begin{tabular}[c]{@{}l@{}}Full page - without\\ font fine-tuning\end{tabular}}} & 19.84\% & 50.75\% \\
\addlinespace[0.3cm]
\multicolumn{1}{l}{\textbf{\begin{tabular}[c]{@{}l@{}}Full page - with\\ font fine-tuning\end{tabular}}}    & 15.55\% & 40.80\% \\
\addlinespace[0.3cm]
\multicolumn{1}{l}{\textbf{\begin{tabular}[c]{@{}l@{}}Utilizing layout\\ detection\end{tabular}}} & 8.93\%                                 & 32.75\%                                \\
\addlinespace[0.3cm]
\multicolumn{1}{l}{\textbf{Final score}}                & 5.56\%                                 & 24.11\%                                \\
\botrule
\end{tabular*}
\end{table}

\subsubsection{Experiment: Evaluation of the OCR on documents from 1868}
In order to find out to what degree our solution is capable of processing older documents, we tested it in \textit{Schematismus} documents from 1868. These differ slightly from the 1910 version, which had been used to train and optimize both layout detection and OCR. Following the surprisingly good layout detection results presented in Table \ref{tab:evaluation_metrics_1868}, the two randomly selected documents from 1868 were transcribed manually to provide ground-truth for measuring OCR performance. In order to establish a baseline, the entire pages were first fed into Tesseract and then the average CER and WER were calculated. Then, layout detection was used to segment the pages into individual snippets that were then fed into Tesseract one at a time. Again, this process resulted in an average CER and WER. It is important to note that the randomly selected documents contain relatively simple text, which may explain the positive results presented in the following table. It is likely that the slight increase in word error rate is caused by the layout detection model sometimes splitting individual paragraphs into two paragraphs.

\begin{table}[ht]
    \caption[]{This table shows the average CER and WER when OCR is applied to documents from 1868 with and without layout detection.}
    \label{tab:experiment_1868}
    \begin{tabular*}{\columnwidth}{@{\extracolsep\fill}lcc}
    \toprule
    & \multicolumn{1}{l}{\textbf{Avg. CER}} & \multicolumn{1}{l}{\textbf{Avg. WER}} \\
    \midrule
    \multicolumn{1}{l}{\textbf{Full page}} & 7.42\%                                    & 32.32\%                                   \\
    \addlinespace[0.3cm]
    \multicolumn{1}{l}{\textbf{\begin{tabular}[c]{@{}l@{}}Utilizing layout\\ detection\end{tabular}}}    & 6.59\%                                    & 33.57\%                                   \\
    \botrule
    \end{tabular*}
\end{table}

\section{Discussion}\label{chap:discussion}
It is evident from the results presented in chapter \ref{chap:evaluation} that the use of a custom-developed layout detection model to segment \textit{Schematismus}-style documents together with a Tesseract model fine-tuned to a custom font designed to be as close to the original as possible significantly improved the quality of the extracted text. However, it should be noted that, due to the small sample size, especially for the OCR-evaluation test set, these results might not represent the full picture. The gain that combined layout detection and text extraction provides may be larger than metrics alone can express. That is due to two different reasons.\\
As each layout element is segmented by bounding boxes, it is known where the blocks are located within a document coordinate-wise. 
It allows the reordering of the extracted texts so that they correspond to the reading flow of the document. This is essential when extracting text from a document with columns.\\
The second reason is that due to the classification of the individual bounding boxes it is possible to immediately tell the class of a structure element. Thus, it is possible, for example, to extract only headings and paragraphs from a document. Moreover, this makes it easier to match paragraphs with individual headings, or to get all enclosed structure elements within curly brackets.\\
As for the Tesseract OCR model, the custom designed font should be improved, to make text extraction even better. Due to the fact that the current version of the font does not include certain characters such as "č" or "ň" in its unichar set, detection of these kinds of letters is not possible, resulting in errors.

\subsection{Error Analysis}\label{sec:error_analysis}
\begin{figure}[ht]
    \centering
        \includegraphics[width=\columnwidth]{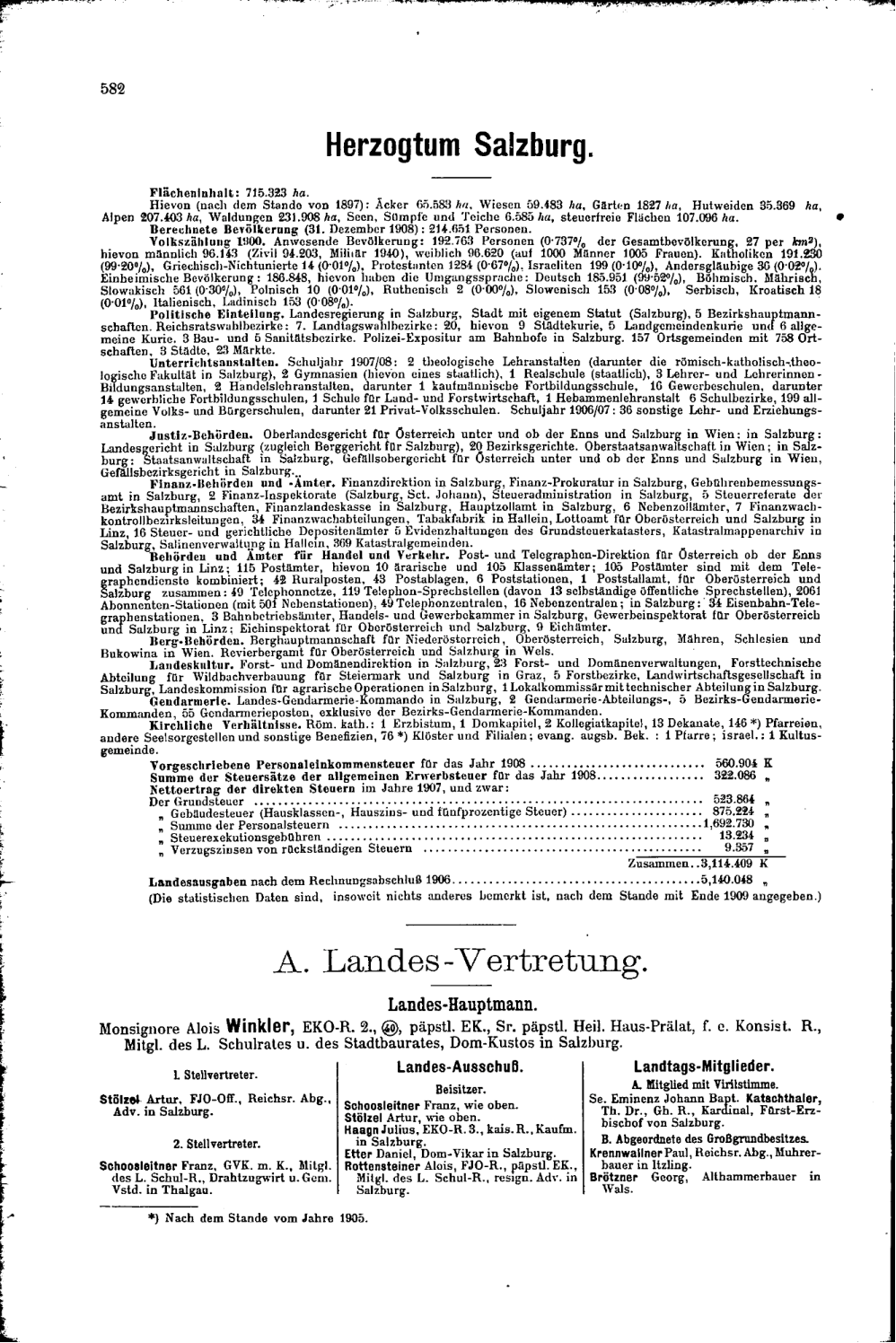}
    \caption[Illustration of poorly performing document.]{This figure illustrates a document with index 4  of the evaluation set, which had the poorest layout-detection performance.}
    \label{fig:error_analysis_4_6}
\end{figure}

According to Table \ref{tab:summary_classification_scores}, documents with the indices four and six performed quite poorly compared to others. As an example, the document with index four is shown in Figure \ref{fig:error_analysis_4_6}. Clearly, this document differs from the typical three-column \textit{Schematismus}-style document.
Aside from the general layout, a key difference is the indentation of each paragraph. Although these aspects were considered during the generation of synthetic documents, resulting in a separate class "BigParagraphs", the generated structures do not appear to be as similar to the originals as intended, based on layout detection results. Due to the relatively small number of examples of this type of document in the training set used for fine-tuning, we could not observe any improvement. 
Specifically, only two pages containing "BigParagraphs" were included in the fine-tuning training set, which appears to be too few for the model to effectively learn this class. 
Therefore, to improve performance on these types of \textit{Schematismus} documents, more pages similar to those in Figure \ref{fig:error_analysis_4_6} must be manually annotated and added to the fine-tuning training set.

\section{Conclusion}
The focus of our study was the analysis of historical state manuals and their subcategories, particularly Central European \textit{Schematismen}, which were widely used throughout Europe during the 19\ts{th} century. These documents contain lists of individuals or institutions working in a particular field and were published by government agencies, churches, or other organizations. As such documents are widely available in PDF format, with either no plain text available or plain text in insufficient quality, it was investigated how OCR accuracy could be improved by segmenting these documents into individual blocks based on a layout detection model trained on a large number of synthetically generated documents and a few original documents, before further processing it with OCR software adapted to historical fonts via finetuning.



In conclusion, we could show how OCR accuracy can be significantly improved by splitting individual document pages into their layout elements as a preprocessing step. Additionally, it has been shown that fine-tuning Tesseract with a custom font results in performance improvement. In comparison with the performance of an out-of-the-box Tesseract Model for OCR on an entire page of the \textit{Schematismus}, the results indicated that segmenting and splitting individual document pages into their layout elements with a deep learning convolutional neural network resulted in up to 71.98 percent better OCR accuracy.

Moreover, the ability to generate a large number of realistic looking, fully annotated, synthetic \textit{Schematismus}-style documents allows access to a large amount of training data. Using this data to train a deep learning neural network, the resulting model can not only be used to significantly speed up the manual annotation process of original document pages through inference, but also act as a pre-trained model for further fine-tuning. In this manner, a layout-detection model can be developed with only a small set of original document pages as a training set. 

The procedure we developed therefore represents a crucial step toward a significantly improved analysis of printed historical documents produced in the larger context of the long 19\ts{th} century, particularly as we show how each of the two steps can be further adapted to the specific needs, requirements and challenges met by fellow researchers. \\
However, we expect that both, layout detection as well as optical character recognition, can be further optimised for even better performance on historical documents. Layout detection may benefit from increasing the training dataset, not only in terms of the number of pages but also by including a wider variety of documents. In other words, by generating documents that are visually similar to much older, in our case \textit{Schematismus}-style documents produced in the first half of the 19\ts{th} century, when a different layout was used, and including these in the training set, a more generalized and robust model may be achieved.\\
Further, domain knowledge can be put to use to enhance text extraction. Considering that most of the printed text in \textit{Schematismus}-style documents consists of abbreviations that are listed and described on specific pages within these documents, this information can be utilized to build a custom spell-checking algorithm to correct errors in the text extraction process. Additionally, it would be of interest to explore whether the methods used in this paper can be applied to other types of historical documents.\\

Our approach offers a viable solution to a number of common problems in dealing with retro-digitised historical texts in historical and humanities research contexts. In this work, it was first shown that the breakdown of the OCR problem, and its solution in several sub-steps, is very promising. 


\bibliography{reference}

\begin{biography}{{\includegraphics[width=77pt,height=77pt,clip,keepaspectratio]{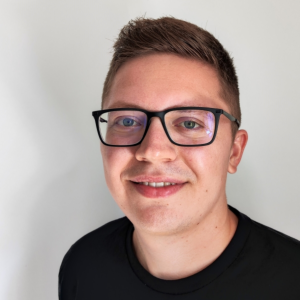}}}{\author{David Fleischhacker} is a computer science master's student at the Technical University of Graz. His interests lie in the field of machine learning driven computer vision, in particular the field of object detection and pattern recognition within images.
\\
\\}
\end{biography}

\begin{biography}{{\includegraphics[width=77pt,height=77pt,clip,keepaspectratio]{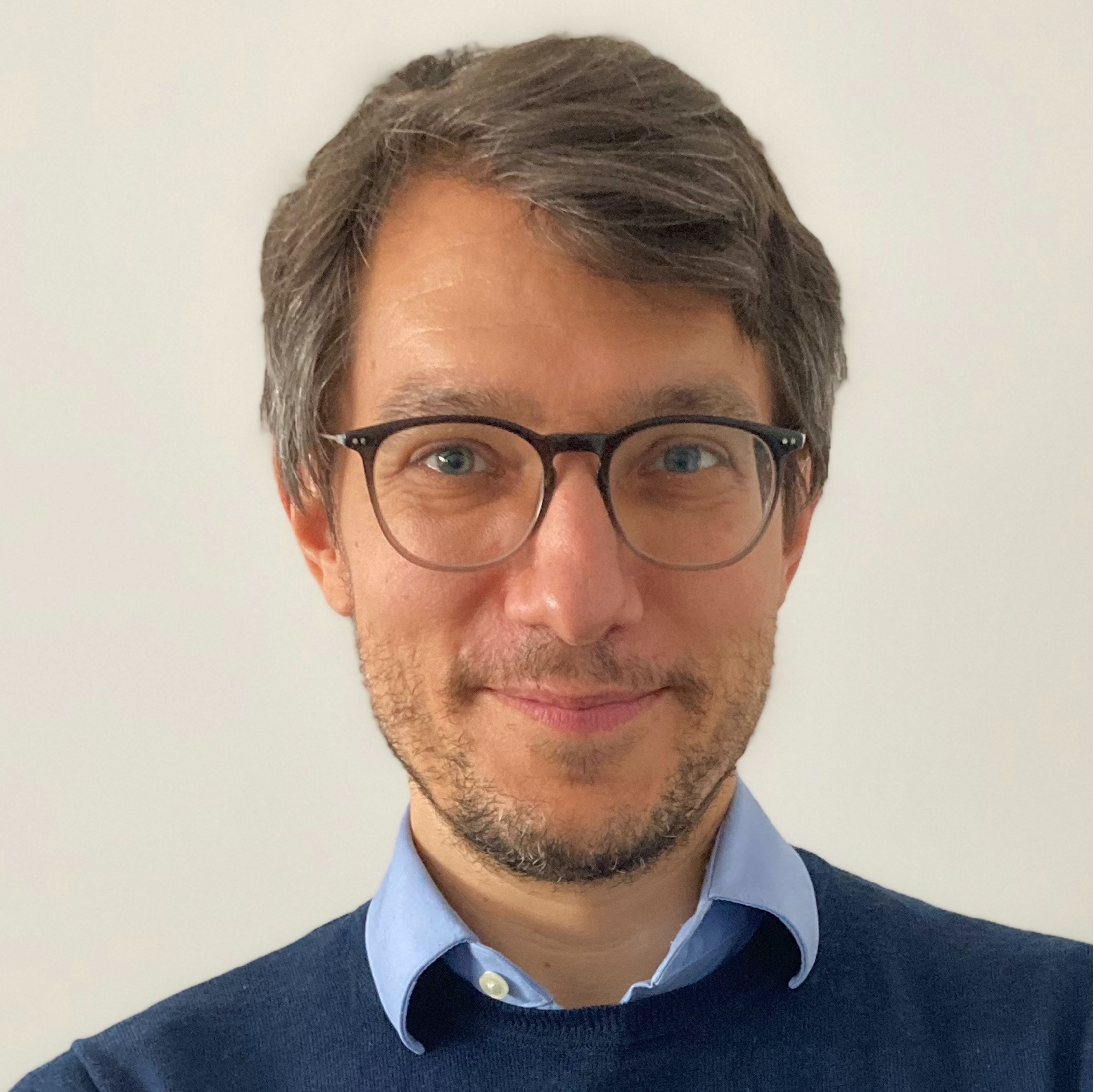}}}{\author{Wolfgang G\"oderle} is a digital historian at the History Department of the University of Graz. His research interests encompass Habsburg Central Europe in the long 19th century, environmental history, and history of science and knowledge, especially large-scale knowledge bases produced by imperial administrations, such as state manuals, cadastral maps and survey. He is particularly interested in using different machine learning-driven approaches to make these primary sources available to research.}
\end{biography}

\begin{biography}{{\includegraphics[width=77pt,height=77pt,clip,keepaspectratio]{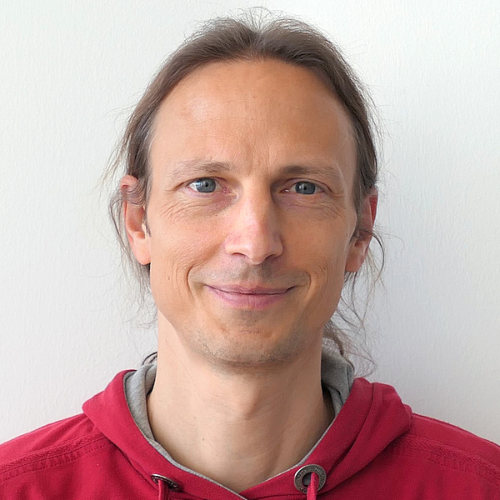}}}{\author{Roman Kern} is a computer scientist and works as Ass.Prof. at the Institute for Interactive Systems and Data Science at the Technical University of Graz. Additionally, he serves as the Chief Scientific Officer at the Know-Center research centre (competence centre for trustworthy AI).
He was awarded his Ph.D. by the Graz University of Technology. 
His research interests include natural language processing and machine learning, and a focus on causal data science.
He applies these methods to achieve trustworthy AI in fields like digital libraries, intelligent transportation systems, and smart production.}
\end{biography}

\end{document}